%% file: main.tex
\documentclass[pdflatex,sn-nature]{sn-jnl}%

\usepackage{graphicx}%
\usepackage{multirow}%
\usepackage{amsmath,amssymb,amsfonts}%
\usepackage{amsthm}%
\usepackage{mathrsfs}%
\usepackage[title]{appendix}%
\usepackage{xcolor}%
\usepackage{textcomp}%
\usepackage{manyfoot}%
\usepackage{booktabs}%
\usepackage{algorithm}%
\usepackage{algorithmicx}%
\usepackage{algpseudocode}%
\usepackage{listings}%
\usepackage{tabularx}
\usepackage{placeins}%
\usepackage{floatpag}
\floatpagestyle{empty}

\begin{document}

\title[Article Title]{Synthesizing symbolic models of human and machine reasoning at scale}

\author*[1,2]{\fnm{Solim} \sur{LeGris}}\email{solim.legris@nyu.edu}

\author[3]{\fnm{Philipp} \sur{Mondorf}}

\author[6]{\fnm{Jinran} \sur{Jin}}

\author[5]{\fnm{Akshay K.} \sur{Jagadish}}

\author[2,4]{\fnm{Brenden M.} \sur{Lake}}

\author[1,6]{\fnm{Todd M.} \sur{Gureckis}}

\affil[1]{\orgdiv{Department of Psychology}, \orgname{New York University}}

\affil[2]{\orgdiv{Department of Psychology}, \orgname{Princeton University}}

\affil[3]{\orgdiv{Center for Information and Language Processing}, \orgname{LMU Munich}}

\affil[4]{\orgdiv{Department of Computer Science}, \orgname{Princeton University}}

\affil[5]{\orgdiv{Princeton AI Lab}, \orgname{Princeton University}}

\affil[6]{\orgdiv{Center for Data Science}, \orgname{New York University}}

\abstract{Symbolic models offer a powerful account of cognition, but their scope has historically been limited by the need for hand-engineering and computational tractability.
We show how advances in artificial intelligence (AI) can scale symbolic modeling and help overcome these engineering limitations. We introduce Behavior-Conditioned Symbolic Synthesis (BCSS)---an approach for automated search of interpretable, deterministic symbolic programs that verifiably reproduce behavior---and apply it to model both human and machine behavior across hundreds of  challenging reasoning problems. Analyzing these programs reveals that both people and machines tend to oversimplify, adopt rules that are partial explanations for their observations, and misinfer dependencies between problem features. These shared signatures are modulated by available resources, with reasoning effort predicting the extent to which machines simplify and explain their observations.
Together, these results point to underlying mechanisms reminiscent of satisficing in both people and machines, and illustrate how BCSS offers a general approach for interpreting and comparing rule-based reasoning of humans and black-box AI systems.}

\maketitle

Symbolic models of cognition provide a long-standing and influential framework for understanding the human mind  \citep{feldmanMinimizationBooleanComplexity2000,fodor1975language,newellPhysicalSymbolSystems1980,braine1998mental,goodman2008rational,slomanCausalModelsHow2005,tenenbaumHowGrowMind2011,lakeHumanlevelConceptLearning2015,yangOneModelLearning2022,ruleSymbolicMetaprogramSearch2024}.
Beyond theoretical arguments that the mind is inherently symbolic and rule-like~\cite{marcus2001algebraic, fodor1975language,newellPhysicalSymbolSystems1980,tenenbaumHowGrowMind2011}, these models offer an alternative to contemporary deep neural networks and foundation models~\cite{lakeHumanlikeSystematicGeneralization2023,webbEmergentAnalogicalReasoning2023,binzFoundationModelPredict2025} that is typically more interpretable and explicit in its representation.
Historically, a key obstacle to constructing symbolic models has been the need for extensive hand-engineering of these representations \citep{mcclellandLettingStructureEmerge2010a,griffithsWhitherSymbolsEra2026}, which can include specifying underlying symbolic primitives and their compositions \citep{tenenbaumRulesSimilarityConcept2000,goodman2008rational,piantadosiLogicalPrimitivesThought2016,lakePeopleInferRecursive2020,sable-meyerLanguageThoughtMental2022,ruleSymbolicMetaprogramSearch2024,zhouCompositionalDiversityVisual2024}.
These demands are tractable in restricted experimental tasks where the relevant hypotheses can be expressed compactly and plausible strategies are approximately enumerable~\cite{goodman2008rational}, but have posed substantial challenges in more open-ended domains.

Here, we leverage recent advances in artificial intelligence (AI) to automatically synthesize symbolic models at scale in an open-ended domain that is challenging for both people and machines. In particular, we exploit the emergence of strong programming abilities in LLMs \citep{chenEvaluatingLargeLanguage2021,austinProgramSynthesisLarge2021,deepseek-aiDeepSeekCoderV2BreakingBarrier2024} to address the issue of combinatorial search within vast discrete program spaces. We developed Behavior-Conditioned Symbolic Synthesis (BCSS), an approach that uses LLMs to automatically search for programs that verifiably reproduce human and machine behavior on problem-solving tasks.

To demonstrate the potential of this approach, we use BCSS to analyze the Abstraction and Reasoning Corpus (ARC), an open-ended set of abstract rule induction tasks spanning a wide range of concepts \citep{chollet2019measureintelligence}.
Rather than focus on correct responses, we focus on modeling patterns of errors in a recently published large dataset of humans solving these problems~\cite{legrisComprehensiveBehavioralDataset2025}, alongside a comparison of errors made by state-of-the-art open weight large reasoning models (LRMs). Our hypothesis was that \textit{incorrect} responses can serve as a rich window into underlying cognitive processes. However, the diversity of errors in a dataset of this size (over 1,700 participants and four model configurations, yielding 6,261 unique human and 4,166 configuration-unique machine errors) means that hand-engineering solutions is impossibly costly.

Our application of BCSS to ARC problems reveals three main insights into human and machine reasoning. First, approximately half of both human and machine incorrect responses we considered are captured by BCSS, suggesting that human and LRM errors are often clustered in regions of program space close to correct reference solutions of ARC problems. Second, analysis of these reverse-engineered programs reveals shared error signatures: both people and LRMs simplify relative to the target rule, find rules that only explain some of the observed evidence, and infer incorrect dependencies between problem features. Third, these signatures are modulated by available resources, with reasoning effort predicting the extent of simplification, partial explanation and reliance on misinferred feature dependencies. More broadly, BCSS programs support investigating patterns of generalization, can expose representational biases and provide bottom-up insights into cognitive mechanisms supporting reasoning. Unlike previous methods, each synthesized symbolic artifact provides a detailed explanation of individual-level behavior, enhancing interpretability of human or machine behavior. We use this fact to expose similarities and differences in abstract reasoning between people and machines using a shared symbolic representational space. We conclude by discussing potential applications of this approach, for example to study mathematical reasoning errors, and the broader implications it has for models of cognition.

\section*{Behavior-Conditioned Symbolic Synthesis}

\begin{figure}[p]
  \centering
  \includegraphics[height=0.85\textheight,keepaspectratio]{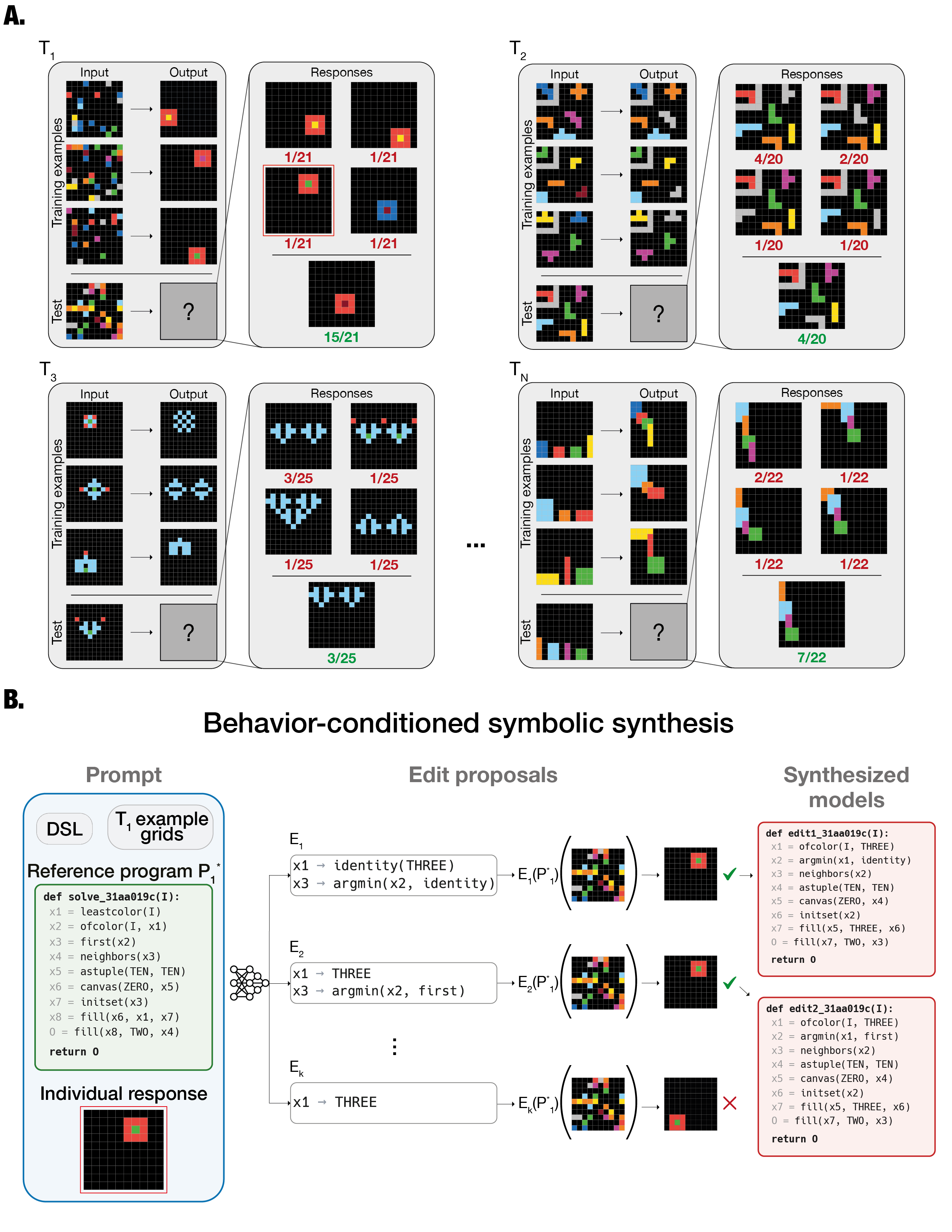}
  \caption{\textbf{Reverse-engineering individual rules in a complex cognitive domain using programs}. (\textbf{A}): Schematic task corpus and associated behavioral data. Each of the four representative ARC tasks $T_i$ presents an agent with a small set of input--output training examples from which a hidden transformation rule must be inferred. From a test input, the agent must then construct the corresponding output grid (hidden with a grey square and ``?'' for illustrative purposes). We use H-ARC, a dataset of responses from over 1,700 participants to the 800 publicly available ARC-AGI-1 tasks. On the right of each task, we show 4 of the most frequent incorrect responses, and the correct response at the bottom. Under each grid, we show the associated response frequency (red for incorrect, green for correct). (\textbf{B}): Behavior-conditioned symbolic synthesis (BCSS). For each unique incorrect response, a large language model is prompted with a full specification of the domain-specific language, task grids, a reference program $P^*_i$ implementing the target rule for task $T_i$, and the observed incorrect response (here, from a human participant). It proposes $k$ independent sets $E_1,\dots,E_k$ of at most $j$ line edits to that program. The schematic edit proposals $E_l$ show LLM-sampled proposals which are then applied to $P^*_i$. Each proposed set of line edits is applied programmatically to the reference program to obtain a new program $E_l(P^*_i)$. A custom parser rewrites the programs such that only causally relevant code is retained (i.e., unreferenced lines of code are removed). Each new program $E_l(P^*_i)$ is then executed with the test input as argument. Programs whose output reproduces the observed response are kept (green checkmark); all others are discarded (red cross). Because a response is often reachable by more than one unique program, BCSS returns a collection of programs per response rather than a single model. The individual response shown here is also shown surrounded by a red square to the right of task $T_1$ in panel A. The constant \texttt{THREE} represents the color green.}\label{fig:main-fig}
\end{figure}

Each ARC task presents a small number of input--output pairs of colored grids (typically two to five) that demonstrate a hidden transformation rule. The solver must infer the rule and apply it to a new test input by constructing an output grid cell-by-cell (Figure~\ref{fig:main-fig}A). Each task is a challenging problem of program induction given the limited number of examples and the effectively unbounded space of possible transformations. Reasoners can go wrong in many ways: adopting a rule that fits only some of the examples, attending to the wrong feature of the objects involved, or identifying the correct rule but carrying out only part of it.

The four subsets of participant-level response examples in Figure~\ref{fig:main-fig}A illustrate that incorrect responses often reflect systematic patterns of inference rather than random or defaulting behavior (incorrect solutions cluster near the correct one shown at the bottom). Similarly to prior work in diagnosing errors in children's mathematical reasoning \citep{brownDiagnosticModelsProcedural1978,vanlehnMindBugsOrigins1990}, we hypothesized that such responses may correspond to constrained deviations from a normative standard. Additionally, we  assume that the rule-like inferences of people and machines in this domain are best understood as programs \citep{acquavivaCommunicatingNaturalPrograms2022}.

We operationalize this hypothesis using simple but expressive programs manipulating grids and implemented in a symbolic domain-specific language (DSL). The DSL we use was developed for expressing programmatic solutions to ARC and includes correct, human-written reference programs for the 400 \verb|training| tasks of ARC-AGI-1 \citep{hodelAddressingAbstractionReasoning2024}. Although BCSS generates a symbolic representation of the underlying algorithm a person or AI system may have applied to produce their response, it is not by itself a process-level description of how a program was inferred. Future work using our approach while leveraging more detailed behavioral traces from the reasoning process (e.g., think aloud protocols) could help expand our understanding of \textit{how} people infer these rules. Here, we instead focus on characterizing the space of final programs applied to produce responses in people and machines.

Specifically, for a given individual response, our method searches for programs representing output behavior by sampling constrained sets of deviations from a reference program. As illustrated in Figure~\ref{fig:main-fig}B, given a reference program $P^*_i$ implementing the target rule for a task $T_i$, an LLM proposes $k$ independent sets of at most $j$ line edits $E_1,\dots,E_k$ to that program. Importantly, proposal sets are sampled conditioned on single human or machine behavioral responses. Each resulting candidate program $E_l(P^*_i)$ is then executed and retained only if it reproduces the response the proposal LLM was conditioned on. Thus, the LLM guides search over a discrete program space by supplying a proposal distribution for plausible deviations implemented as line edits, whereas program execution provides a hard verification mechanism.

\clearpage
\enlargethispage{35mm}
\clearpage

\section{Results}\label{sec:results}

We leverage the H-ARC dataset, a large corpus of human responses from over $1{,}700$ participants to the full set of 800 publicly available ARC-AGI-1 tasks \citep{legrisComprehensiveBehavioralDataset2025}. H-ARC contains up to three attempts per task from each participant, with correct/incorrect feedback after each attempt. We filtered H-ARC to narrow down the dataset to responses more likely to arise from conceptual rather than procedural mistakes (\S\ref{sec:methods-data} for more details). The final filtered H-ARC dataset contains 6,261 unique incorrect outputs across 723 tasks and 1{,}387 participants. For the 400 \verb|evaluation| tasks which did not have a human-written solver, we successfully synthesized reference programs for over 90\% of the tasks (see \S\ref{sec:synthetic-solvers} for more details), and conditioned error program synthesis using those programs. Unless specified, we report human results aggregated over modeled attempts from the \verb|training| and \verb|evaluation| splits of ARC-AGI-1. Wherever substantial divergence was observed, we report results separately. To compare human and machine errors, we additionally evaluate a range of large reasoning models (LRMs) on the 400 \verb|training| tasks of ARC-AGI-1 (see \S\ref{sec:llm-errors} for more details). Because each response is typically explained by several candidate programs, and tasks differ widely in how many errors they elicit, observations are unevenly distributed across programs, errors and tasks. We therefore report estimates from hierarchical Bayesian regressions \citep{mcelreath_statistical_2020}. This approach allows us to pool information within each error and across errors within each task or group, so that heavily sampled errors and tasks do not dominate the results (\S\ref{sec:methods-stats}). Accordingly, we report posterior means or medians with 94\% highest-density intervals (HDIs).

\subsection{Synthesized programs capture human errors}\label{sec:results-coverage}

After verification and filtering (see \S\ref{sec:methods-pipeline} for more details), we obtained 8,246 total programs matching individual responses to tasks in both the \verb|training| and \verb|evaluation| splits of ARC-AGI-1. These programs account for 52.5\% of non-trivial incorrect attempts on the \verb|training| split, and 46.2\% on the \verb|evaluation| split (49.1\%, 3,636/7,402 pooled; Supplementary Table~\ref{tab:sup_coverage} for the full per-split breakdown). Given the edit constraints on the synthesizer, our results demonstrate that a large proportion of incorrect human responses can be modeled by programs a few line edits away (M=2.84, SD=2.04, \S\ref{sec:methods-pipeline}) from reference programs implementing the target rule for each task. Further, among successfully modeled incorrect responses, we recover a median of 2 distinct programs per unique error grid (IQR: [1, 4]). This many-to-one correspondence is illustrated in Figure~\ref{fig:main-fig}B with two distinct, but successful edit sets. The first replaces the \texttt{leastcolor(I)} primitive operation with a hardcoded color, whereas the second additionally substitutes a selection operator. Both yield programs whose output reproduces the same participant response, and both are retained.

We compared BCSS against an exhaustive search over all single-line rewrites of the \verb|training| reference programs (\S\ref{sec:methods-pipeline}). Despite enumerating over 16M edits, this baseline captures only 25.8\% (899/$3{,}485$) of erroneous attempts---roughly half of what BCSS achieves on the same data. With only 15 synthesized samples per response, BCSS still recovers 89\% of the exhaustively searched programs that match human responses. Because the number of candidates grows combinatorially with edit depth, exhaustive search is intractable at the edit distances where many human errors lie (see \S\ref{sec:validation} for validation of BCSS using manually verified synthetic errors).

Under the view of ARC as program induction, our results suggest that people often infer incorrect programs that are structurally similar to the program implementing the target rule for a given task, providing support for our initial hypothesis that incorrect responses in this open-ended domain are not merely idiosyncratic failures.

\subsection{Inferred programs are systematically simpler}\label{sec:results-simplicity}

\begin{figure}[p]
  \centering
  \includegraphics[height=0.85\textheight,keepaspectratio]{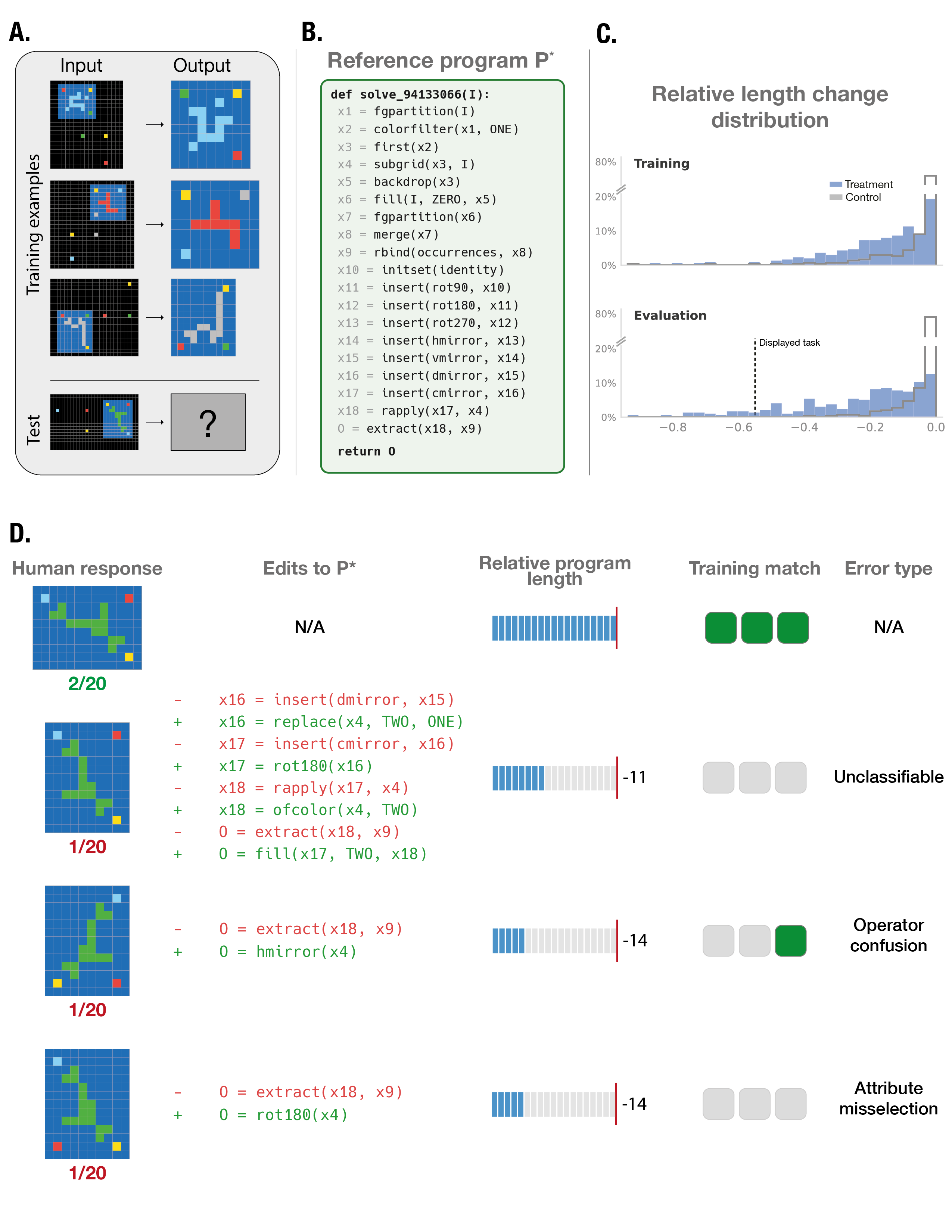}
  \caption{\textbf{Synthesized programs suggest people often infer rules simpler than the target rule for a given task}. (\textbf{A}): A representative ARC task (\texttt{94133066}). Three training examples demonstrate the target rule---return the contents of the framed sub-grid under whichever of its eight rotations and reflections is consistent with the isolated colored cells scattered around it---followed by the test input for which participants construct an output grid (hidden with a grey square and ``?'' for illustrative purposes). (\textbf{B}): The reference program $P^*$ implementing the target rule for this task. Because this task belongs to the \texttt{evaluation} split, $P^*$ is itself model-synthesized rather than human-written. (\textbf{C}): Distribution of relative program length change, $(L(p) - L(P^*))/L(P^*)$, over all synthesized programs, shown separately for the \texttt{training} and \texttt{evaluation} splits of ARC-AGI-1. Treatment programs (blue) are those conditioned on incorrect human responses; control programs (grey) are produced by an identical procedure conditioned instead on the correct output grid. Because edits replace lines in-place and causally irrelevant code is subsequently removed, synthesized programs can either shorten, or remain the same, but never lengthen. Therefore, the control condition controls for the shortening bias attributable to the procedure itself, against which the leftward displacement of the treatment distribution should be read. The dashed line marks the median shortening of modeled responses for the task displayed in this figure. (\textbf{D}): The correct response (top row) and the three of the most frequent incorrect responses to that task, with response frequencies below each grid (green, correct; red, incorrect). For each, we show the diff with $P^*$ after applying BCSS, the length of the resulting program relative to $P^*$ (blue, retained lines; gray, removed lines; red rule, reference length), which training examples the program reproduces (green, reproduced), and its error category. A single edit can account for a large simplification: for example, replacing the final \texttt{extract} over the eight candidate transforms by one fixed transform renders fourteen lines of the reference program inconsequential, and they are removed. Thus, the simpler resulting program hardcodes a transform instead of exhaustively searching over possible transformations.}\label{fig:simplicity-fig}
\end{figure}

Human learners intrinsically value simple representations over more complex ones~\citep{chaterSimplicityUnifyingPrinciple2003,feldmanSimplicityPrincipleHuman2003,kendallComplexityFormingMental2024}, and this fact is sometimes an essential component of models posited to capture human behavior \citep{goodman2008rational}. Here, we examine whether this bias for simple codes is a property of the behavioral outputs which we modeled using BCSS, beyond the shortening bias intrinsic to the BCSS edit procedure. As a test of this idea, we applied the same procedure used for producing error programs using identical reference programs, but instead conditioned the synthesizer on \textit{correct} output grids (see \S\ref{sec:methods-stats} for more details on statistical models). This measures the inherent tendency of BCSS to produce shorter programs independent of the nature of the errors. Because the programs are adapted using line-level edits, we use relative line-count change as a proxy for difference in program complexity with respect to the reference programs. Statistical analyses confirm that DSL programs that capture human responses are credibly simpler than reference programs they are adapted from, on both splits (Figure~\ref{fig:simplicity-fig}C). To obtain these results, we fit a regression predicting average relative line-count change for error programs and control programs. On the \verb|training| split, where each reference program is hand-written by a human, error programs are 19.1\% shorter on average (94\% HDI $[-0.206, -0.176]$), compared with only 5.2\% for baseline (correct-output) programs (94\% HDI $[-0.076, -0.028]$). This corresponds to an additional shortening of 13.9 percentage points attributable to conditioning on human errors (94\% HDI $[-0.164, -0.116]$). The same pattern holds on the \verb|evaluation| split, with the caveat that its reference programs are also model-synthesized rather than hand-written. Error programs from the \verb|evaluation| split are 24.3\% shorter on average (94\% HDI $[-0.259, -0.225]$) versus 2.7\% for baseline programs (94\% HDI $[-0.055, -0.002]$). This corresponds to an additional 21.5 percentage points attributable to conditioning on human errors (94\% HDI $[-0.240, -0.189]$). Panel D of Figure~\ref{fig:simplicity-fig} illustrate a drastic but intuitive example of simplification: the target rule for this task requires searching over the eight rotations and reflections of a framed sub-grid for the one consistent with the surrounding cells, but the programs reproducing the two bottom responses instead use fixed transforms (\texttt{hmirror} and \texttt{rot180}, respectively), rendering fourteen lines of the reference program irrelevant to the final output.

To examine this effect further, we fit an additional Bayesian regression predicting relative line-count change in error programs as a function of reference program line count, empirical task success rate and error categories for \verb|training| split tasks only (see \S\ref{sec:results-taxonomy} for further discussion of error categories; see \S\ref{sec:error-cat-diff-model} for details on the regression). The statistical model finds that reference program complexity modulates the magnitude of relative shortening. For each additional line in a reference program, we find 0.47 percentage points more shortening on average in corresponding error programs (94\% $[-0.0056, -0.0037]$). For example, error programs derived from a 7-line reference program are predicted to be 5.8\% shorter on average, whereas those from a 28-line reference are 15.7\% shorter. This corresponds to a 9.8 percentage-point difference for a four-fold increase in reference program length (Supplementary Figure~\ref{fig:sup_length}A).

Additionally, we found suggestive evidence that conditional on a reference program's line count, errors from tasks with higher empirical success rate are modeled by slightly less relative shortening (94\% HDI $[-0.0213, 0.0666]$). The estimated coefficient suggests that easier tasks may tend to elicit error rules closer to the reference programs in complexity, but the credible interval includes zero so these results remain inconclusive.

Although the procedure is biased toward shortening, our analysis suggests that, beyond this bias, the incorrect rules that people infer are often represented by substantially simpler programs relative to the program implementing the target rule for a given task. These results hint at a bounded cognition account, where people possibly adopt increasingly \textit{simpler} rules when the target rule is costlier to infer or verify.

\clearpage
\enlargethispage{35mm}
\clearpage

\subsection{Inferred programs partially satisfy available evidence}\label{sec:results-partial}

\begin{figure}[p]
  \centering
  \includegraphics[height=0.85\textheight,keepaspectratio]{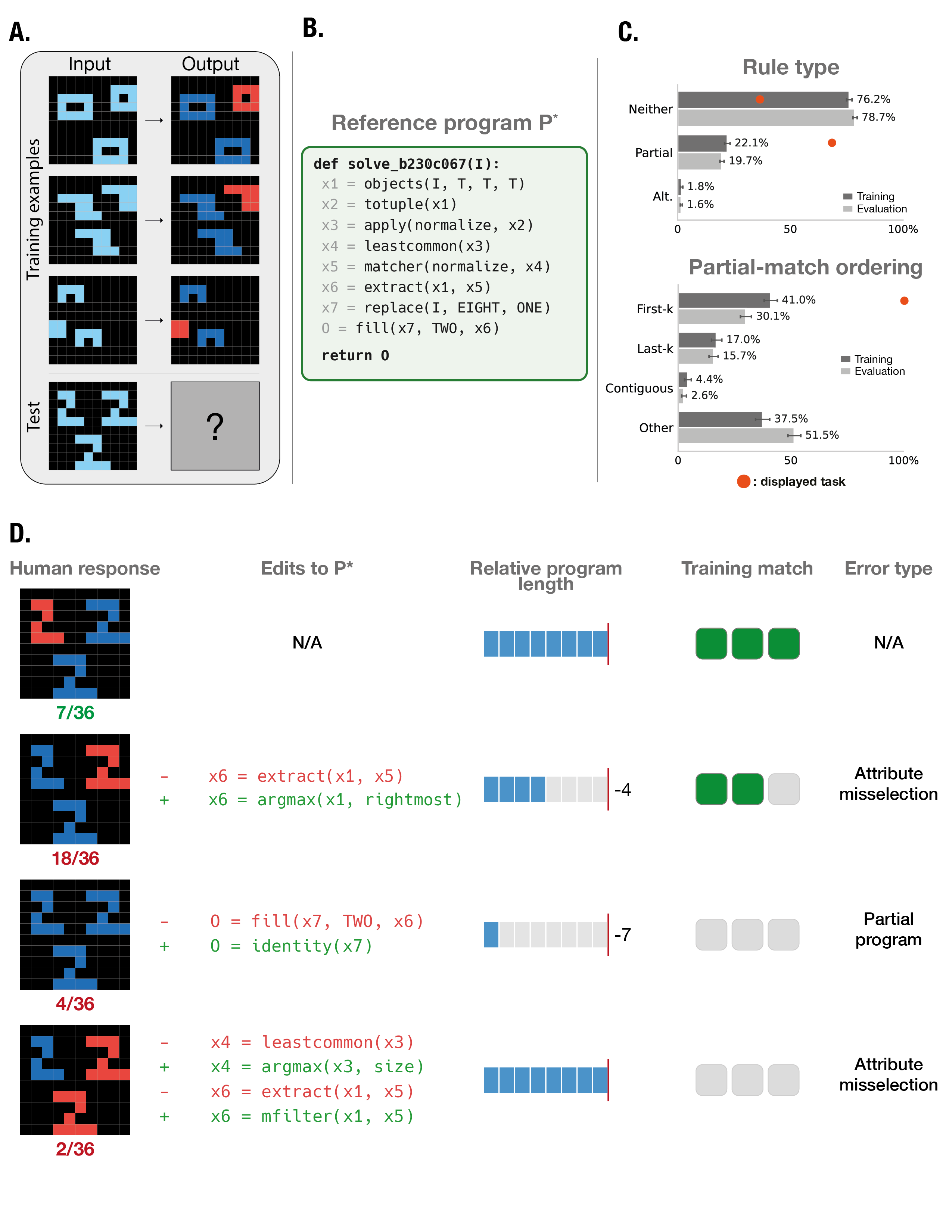}
  \caption{\textbf{Synthesized programs suggest people infer rules that only explain some of the data they observe}. (\textbf{A}): A representative ARC task (\texttt{b230c067}). Three training examples specify the target rule---recolor every shape blue except the least common one, which is recolored red---followed by the test input for which participants construct an output grid (hidden with a grey square and ``?'' for illustrative purposes). (\textbf{B}): A human-written reference program $P^*$ implementing the target rule for this task. (\textbf{C}): Top, prevalence of each rule type among all modeled incorrect attempts. A program is a \emph{partial-match} if it reproduces at least one, but not all, of the training outputs when executed on training inputs; an \emph{alternative rule} if it reproduces all of them yet still fails the test example; and \emph{neither} otherwise. Bottom, conditional on an attempt being explained by a partial-match program, the positions of the training examples it reproduces: the first-$k$ or last-$k$ example(s) shown on screen, a contiguous block of at least two examples that doesn't include the first or last examples, or any other configuration. In both plots, bars give the posterior median attempt-weighted prevalence and error bars the 94\% HDI, shown separately for the \texttt{training} (dark) and \texttt{evaluation} (light) splits of ARC-AGI-1; orange dots give the corresponding proportion for the task shown in panel A. (\textbf{D}): The correct response (top row) and three of the most frequent incorrect responses to that task, with response frequencies below each grid (green, correct; red, incorrect). For each, we show the diff with $P^*$ after applying BCSS, the length of the resulting program relative to $P^*$ (blue, retained lines; gray, removed lines; red rule, reference length), which training examples the program reproduces (green, reproduced), and its error category. The most frequent incorrect response is captured by a program that selects the rightmost object rather than the least common one, a rule that satisfies the first two training examples but not the third. }\label{fig:partial-match}
\end{figure}

Because we represent the rules that people inferred as programs, we can verify how these programs behave on the training examples that participants observed. For each program that reproduces a human response and its associated task, we executed it on each of the training input grids to determine whether it produced any correct output grids. To aggregate over multiple programs fitting each error, we fit a Bayesian categorical regression predicting the extent to which errors are modeled by programs that reproduce none, some or all demonstrations seen by participants (see \S\ref{sec:rule-type-model} for more details on the statistical model). Overall, we find that 22.6\% (94\% HDI $[21.9\%, 23.4\%]$) of modeled incorrect attempts correspond to programs consistent with at least one training example, with no major differences between splits (see Supplementary Table~\ref{tab:sup_ruletypes} for per-split results). This subset of programs is split between two rule types: partial-match and alternative rule (Figure~\ref{fig:partial-match}C, top). Partial-match programs correspond to rules where a strict subset of training outputs are correctly generated by the error program and account for 20.9\% of reproduced incorrect attempts (94\% HDI $[20.2\%, 21.7\%]$). Conversely, alternative rules correspond to programs that correctly reproduce all observed training examples and the behavioral response (and thus are arguably equally valid), but differ in their prediction for the test output. These alternative rules captured 1.7\% (94\% HDI $[1.4\%, 2.0\%]$) of modeled attempts. The first incorrect response in Figure~\ref{fig:partial-match}D illustrates a partial-match rule: the program that captures it selects the rightmost object rather than the least common one, a rule that happens to satisfy the first two training examples but fails on the third.

We further examined synthesized programs for order effects: do partial-match programs suggest order-dependent inference or verification of rules? We labeled each partial-match program by which examples it reproduces: the first or the last $k$ examples shown on screen, a contiguous block of at least two examples, and anything else as ``Other'' (Figure~\ref{fig:partial-match}C, bottom). We then fit a Bayesian categorical regression predicting the extent to which errors tend to be modeled by programs belonging to one of these four categories (see \S\ref{sec:rule-type-model} for more details on the statistical model). On the \verb|training| split, 62.5\% (94\% HDI $[59.4\%, 65.6\%]$) of partial-match errors form a structured subset, dominated by the first-$k$ pattern at 41.0\% (94\% HDI $[38.2\%, 44.0\%]$), followed by 17.0\% (94\% HDI $[14.8\%, 19.4\%]$) matching the last-$k$ and 4.4\% (94\% HDI $[2.9\%, 5.9\%]$) a contiguous block; the remaining 37.5\% (94\% HDI $[34.4\%, 40.6\%]$) match a single interior example or a non-contiguous set. On the \verb|evaluation| split, the structured share is lower at 48.5\% (94\% HDI $[45.6\%, 51.3\%]$), with 30.1\% (94\% HDI $[27.6\%, 32.6\%]$) first-$k$, 15.7\% (94\% HDI $[13.8\%, 17.8\%]$) last-$k$, and 2.6\% (94\% HDI $[1.6\%, 3.7\%]$) contiguous block.

Despite BCSS not having explicit constraints to encourage the types of programs we describe here, these results suggest that people often find rules that are incomplete explanations of the data they observe. We hypothesize that such rules may be the consequence of ordered, sequential rule induction or verification, paired with early stopping. Taken together, these results are consistent with satisficing \citep{simonRationalChoiceStructure1956,simonModelsBoundedRationality1982}: people adopt a rule based on whether it satisfies a ``good enough'' threshold, rather than verifying it against all evidence or expending more cognitive effort to find better, potentially more complex rules that explain all the observed data.

\clearpage
\enlargethispage{35mm}
\clearpage

\input{tables/error_taxonomy.tex}

\subsection{Inferred programs follow common error patterns}\label{sec:results-taxonomy}

Prior work has sought to characterize reasoning error kinds, for example on intelligence tests \citep{carpenterWhatOneIntelligence1990,kundaErrorPatternsRavens2016}. Contrary to success rate, which is a sparse signal indicating differences in \textit{degree}, error analyses afford characterizing differences in \textit{kind}, which is informative about underlying cognitive strategies. Unlike the data we model here, researchers often collect forced-choice responses (e.g., Raven's Progressive Matrices) rather than richer, free-form responses. This implies that the scope of possible error types is predefined by the structure of the incorrect alternative responses of each task. In comparison, we can more richly characterize the space of reasoning errors that people make, without strong constraints on their responses.

\begin{figure}[p]
  \centering
  \includegraphics[width=\textwidth,height=0.85\textheight,keepaspectratio]{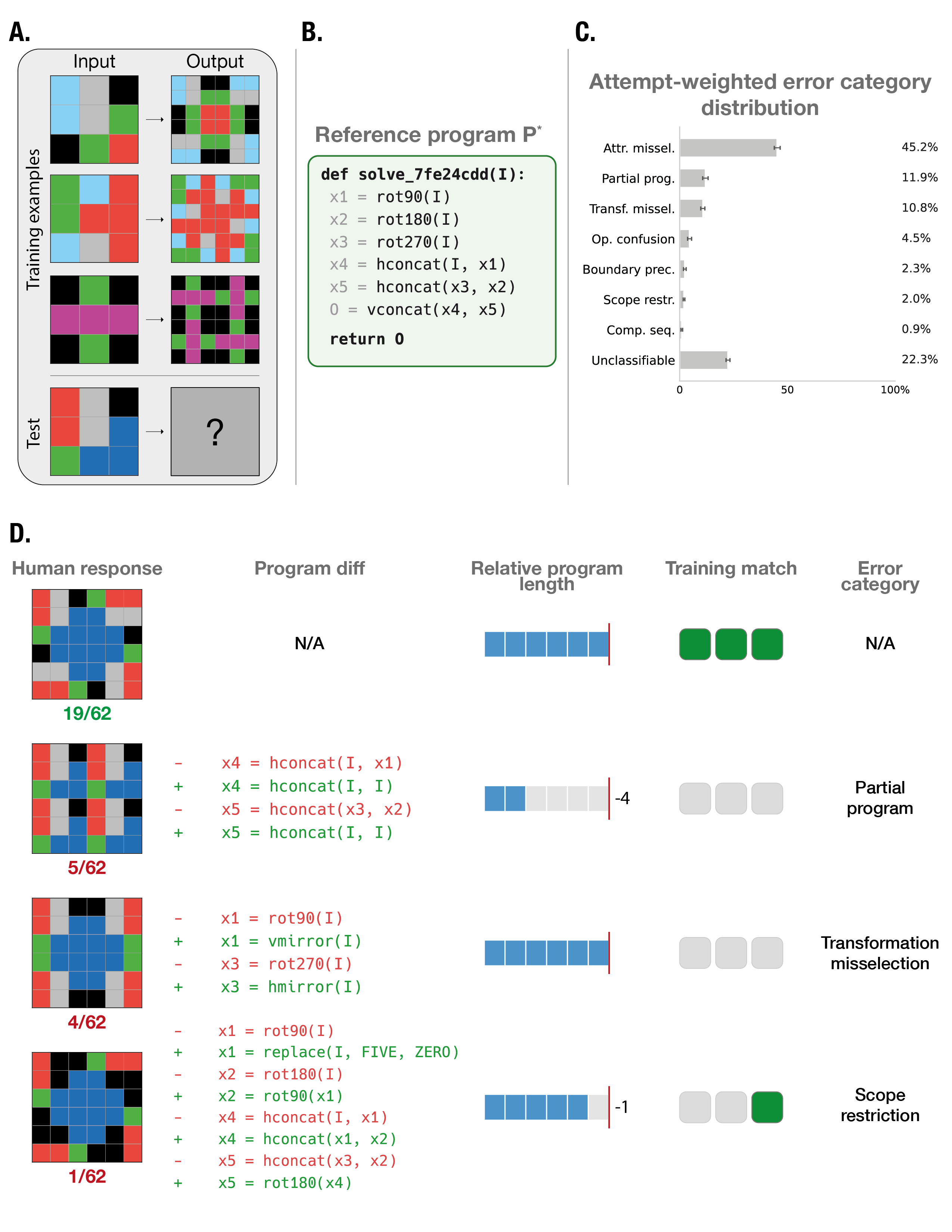}
  \caption{\textbf{Synthesized programs reveal a restricted set of recurring deviations from target rules}. (\textbf{A}): A representative ARC task (\texttt{7fe24cdd}). Three training examples specify the target rule---tile the input into a $2\times2$ arrangement in which each quadrant holds the input rotated by a further 90$^\circ$ clockwise, starting from the input at the top left---followed by the test input for which participants construct an output grid (hidden with a gray square and ``?'' for illustrative purposes). (\textbf{B}): A human-written reference program $P^*$ implementing the target rule for this task. (\textbf{C}): Prevalence of each error category among modeled incorrect attempts. An LLM classifier assigned every program one of seven cognitive categories, given the reference program, the task grids, the synthesized program and its execution trace (Table~\ref{tab:error-taxonomy} for definitions). Bars give the posterior median of the attempt-weighted prevalence derived from a hierarchical Bayesian categorical model, and error bars the 94\% HDI; the categories are mutually exclusive. Classification was performed on programs fitting responses to \texttt{training} split tasks only. (\textbf{D}): The correct response (top row) and three of the most frequent modeled incorrect responses to the task in panel A, with the response frequency below (green, correct; red, incorrect). For each, we show the diff with $P^*$ after applying BCSS, the length of the resulting program relative to $P^*$ (blue, retained lines; gray, removed lines; red rule, reference length), which training examples the program reproduces when executed on their inputs (green, reproduced), and its error category. The programs that reproduce these three incorrect responses diverge from $P^*$ in different ways: omitting the rotations and tiling the input four times, carrying out reflections instead of rotations, or correctly carrying out the rule on a restricted portion of the input.}
  \label{fig:error-classification}
\end{figure}

To classify types of errors, we devised a novel error taxonomy comprised of seven cognitive categories and a catch-all ``unclassifiable'' category (see Table~\ref{tab:error-taxonomy}). For each valid synthesized program of tasks in the \verb|training| split, we sampled labels using an LLM classifier. We then aggregated program-level classifications at the attempt level using a hierarchical Bayesian model (see \S\ref{sec:methods-classification} for more details). Overall, the model finds a dominant category of errors: \textit{attribute misselection} at 45.2\% (94\% HDI [43.9, 46.6]) of modeled incorrect attempts (Figure~\ref{fig:error-classification}C). Inspired by \cite{kahnemanRepresentativenessRevisitedAttribute2002}'s work on attribute substitutions in intuitive judgment, we define attribute misselections as errors that arise when a reasoner fails to identify or compute the correct dependency between an attribute and a target parameter. The first incorrect response shown in Figure~\ref{fig:partial-match}D demonstrates an example of such a rule: determining which object to color red (i.e., the target parameter) as a function of which object is odd with respect to the others (i.e., the attribute). Instead, participants often rely on possibly simpler or more salient dependencies. Observing that the top-right object is always red is a correct inference based on the first two examples, but it does not hold in the third example. This error category aims to capture simplifications of this kind, and our results suggest that people often rely on such heuristics.

Additionally, we find that people often infer rules that correspond to \textit{partial programs}, representing 11.9\% (94\% HDI [10.7, 13.2]) of modeled incorrect attempts. Such errors arise when participants correctly applied a rule that captures a subset of the correct steps for solving a task. Finally, we find that 10.8\% (94\% HDI [9.8, 11.7]) of incorrect attempts are transformation misselections, where the participant applied an incorrect transformation to the right parts of the input. The remaining four cognitive categories represent less than 5\% of modeled incorrect attempts, and 22.3\% (94\% HDI [21.5, 23.3]) are determined to be unclassifiable under our taxonomy.

Every modeled response we display is labeled with its modal error category (panel D, last column of Figures~\ref{fig:simplicity-fig}--\ref{fig:error-classification}), illustrating the different ways in which participants can deviate from a single target rule. For task \texttt{7fe24cdd} in Figure~\ref{fig:error-classification}, the rule is to tile the input into a $2\times2$ arrangement of successive $90^\circ$ rotations. In that figure's panel D, the first response omits the rotations and simply tiles the input four times (\textit{partial program}, four lines shorter), the next preserves the construction but substitutes reflections for rotations (\textit{transformation misselection}, same length), and the third applies the construction to a subset of the input (\textit{scope restriction}, one line shorter). As suggested by how these examples differ in shortening, beyond the effect of target rule complexity (\S\ref{sec:results-simplicity}), we observe general quantitative differences in relative shortening across error categories, with localized errors such as boundary precision leaving program length largely unchanged, and partial program errors substantially shorter than the target rule (Supplementary Section~\ref{sup:length}).

Taken together, these results provide evidence that errors on ARC tasks are not idiosyncratic deviations from the target rule. Most modeled incorrect attempts fall into a small set of categories, with attribute misselection as the most commonly occurring error type. Individual categories also tell us how much of the target rule's structure is kept. Localized errors retain most of it, while partial programs and transformation misselections are coarser approximations. Error analyses of BCSS programs can therefore not only capture whether people's failures are rule-like, but also how their inferences diverge from the target rule.

\clearpage
\enlargethispage{35mm}
\clearpage

\subsection{Comparing human and machine errors}\label{sec:llm-errors}

Since the advent of reasoning-tuned LLMs \citep{guoDeepSeekR1IncentivizesReasoning2025,openaiOpenAIO1System2024}, performance on ARC has drastically increased, reaching near-ceiling performance among the most capable models.\footnote{See the ARC Prize leaderboard, \url{https://arcprize.org/leaderboard} (accessed October 2026).} Given observed scaling trends of performance with respect to reasoning effort on these problems, we investigate whether the patterns of results we established in people differ for these models, and how modulating reasoning effort might vary results. We study two reasoning-tuned LLMs (hereafter referred to  as large reasoning models or LRMs), Qwen3.5-27B~\citep{qwen3.5} and GPT-OSS-120B~\citep{openai2025gptoss120bgptoss20bmodel}, both of which generate long chains of thought (CoTs) before producing an answer. Because GPT-OSS-120B allows us to vary how much the model dynamically allocates tokens for CoT before providing an answer (i.e., varying test-time compute), we evaluate it at high, medium, and low reasoning effort. We sample ten grid responses from each model and configuration on each of the 400 \verb|training| tasks in ARC-AGI-1 (\S\ref{sec:methods-llms}).

\begin{figure}[tbp]
  \centering
  \includegraphics[width=1\textwidth]{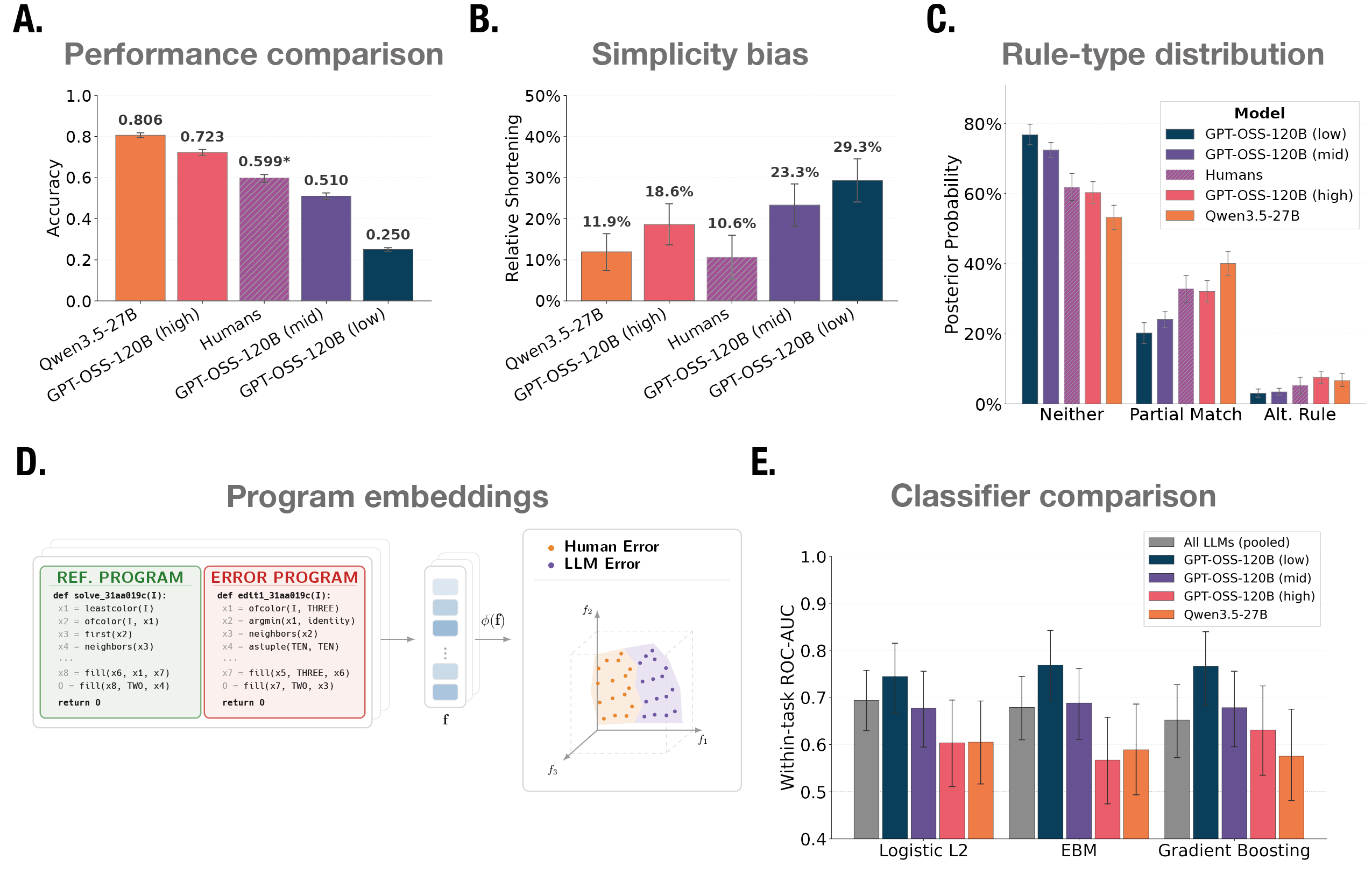}
  \caption{\textbf{Machine errors are systematic and performance-dependent.} (\textbf{A}): Average accuracy across responses of large reasoning models at different reasoning efforts and humans on the \texttt{training} set of ARC-AGI-1. Note that GPT-OSS-120B uses a mixture-of-experts architecture and thus only has 5.1 billion active parameters during inference. (\textbf{B}): Relative shortening of programs synthesized from model errors compared with programs for correct-output controls; all model configurations exhibit a simplicity bias, which is stronger for lower-performing configurations. (\textbf{C}): The proportion of error programs that partially match or fully reproduce the support examples increases with model performance. (\textbf{D}) Schematic of the classification analysis distinguishing human and machine errors using features derived from synthesized error programs. (\textbf{E}) Within-task ROC-AUC of three fitted classifiers provides evidence of varying degrees of discriminability of human and machine errors on held-out tasks, with less performant models better discriminated from people. Classifiers are trained using linearized program-level feature vectors. ROC-AUC is computed among the errors of each of the 56 shared tasks and averaged across tasks, with error bars showing 95\% task-level bootstrap CIs. Panels \textbf{B} and \textbf{C} report results on the subset of 56 tasks where both reasoning language models and people generated incorrect responses (see Appendix~\ref{sup:llm_alltasks} for similar trends across all errors). All machine responses are independently sampled, and we restrict comparisons to human first-attempt responses.}
  \label{fig:llm-error-analysis}
\end{figure}

Figure~\ref{fig:llm-error-analysis}A shows the average accuracy across independent attempts and its standard deviation across runs. Performance varies substantially across models and reasoning efforts. Qwen3.5-27B, with approximately 27 billion active parameters, performs best at 80.63\% accuracy, followed by GPT-OSS-120B, with approximately 5.1 billion active parameters, at 72.28\% under high reasoning effort. Both models outperform the estimated average human accuracy of 59.9\% on a single attempt, as reported by~\citep{legrisComprehensiveBehavioralDataset2025}. Reducing reasoning effort substantially lowers GPT-OSS-120B's performance to 50.96\% at medium effort and 25.02\% at low effort. After filtering (\S\ref{sec:methods-data}), we observe 4{,}166 configuration-unique errors across all configurations, spanning 357 of the 400 tasks.

BCSS captures a large fraction of LRM errors, indicating that many admit systematic programmatic descriptions. After verification and filtering (\S\ref{sec:methods-data}), we obtain 3{,}804 programs that capture 62.8\% (1{,}565/2{,}491) of GPT-OSS-120B incorrect attempts at low reasoning effort, 2{,}760 programs that capture 59.3\% (1{,}031/1{,}738) at medium reasoning effort, and 944 programs that capture 54.4\% (355/652) of its incorrect attempts at high reasoning effort. For Qwen3.5-27B, 823 programs capture 51.5\% (296/575) of its incorrect attempts. For comparisons across models and humans, we restrict subsequent analyses to the 56 tasks on which humans and every model configuration make at least one error, so that comparisons are made on errors in a common set of tasks. We find similar trends hold when considering the full pool of errors across models (see Supplementary Tables~\ref{tab:sup_llm_yield}--\ref{tab:sup_llm_categories}). Finally, we only include first-attempt human errors to match the machine evaluation procedure where responses are independently sampled. 

We investigate whether LRM errors exhibit a similar simplicity bias as observed in people by looking at whether programs synthesized from LRM errors are shorter than those synthesized for the correct-output control group (see \S\ref{sec:results-simplicity} for human results; see \S\ref{sec:simplicity-model} for more details on the regression model). Figure~\ref{fig:llm-error-analysis}B shows the average relative shortening of error programs compared with baseline programs. Two notable patterns emerge. First, all model configurations exhibit a simplicity bias similar to humans, but to a varying degree. For the strongest model we tested, Qwen3.5-27B, its error programs are 11.9\% shorter on average than baseline programs (94\% HDI [$-0.164, -0.073$]). For GPT-OSS-120B, error programs are 18.6\% shorter at high reasoning effort (94\% HDI [$-0.237, -0.136$]), 23.3\% shorter at medium reasoning effort (94\% HDI [$-0.285, -0.182$]), and 29.3\% shorter at low effort (94\% HDI [$-0.346, -0.241$]). Second, the magnitude of relative shortening is correlated with accuracy: programs that represent errors of lower-performing systems are shorter relative to baseline than those of higher-performing systems (Pearson's $r = -0.80$, 94\% HDI [$-0.95, -0.62$], $n=5$; \S\ref{sec:simplicity-model}). Each additional 10 points of accuracy corresponds to 3.0 percentage points less shortening (94\% HDI [$1.9, 4.0$]). These results provide model-based support for our hypothesis that simplification is driven by resource constraints: less capable systems, or those devoting less compute to reasoning, tend to infer simpler rules.

As with people, we also analyze the prevalence of rule types and order effects (\S\ref{sec:results-partial}) within programs synthesized from LRM errors. Figure~\ref{fig:llm-error-analysis}C shows that the proportion of partial-match rules increases with model capacity, and alternative rules are substantially more frequent for the two highest-performing configurations. For GPT-OSS-120B at low reasoning effort, only $20.2\%$ ($94\%$ HDI [$17.2\%, 23.2\%$]) of programs produce partial matches and $3.0\%$ ($94\%$ HDI [$1.9\%, 4.2\%$]) represent alternative rules. For Qwen3.5-27B, these proportions rise to $40.0\%$ ($94\%$ HDI [$36.6\%, 43.5\%$]) and $6.7\%$ ($94\%$ HDI [$4.9\%, 8.7\%$]), respectively (Supplementary Table~\ref{tab:sup_llm_ruletypes} for all values across the full set of errors). Only GPT-OSS-120B at high reasoning effort shows a human-like order effect, with more first-$k$ than last-$k$ partial matches ($36.1\%$, $94\%$ HDI [$30.8\%, 41.4\%$] vs.\ $21.8\%$ [$16.9\%, 26.9\%$]; Supplementary Figure~\ref{fig:sup_llm_matched}B). The other configurations show no consistent preference for explaining the first examples, with GPT-OSS-120B at medium reasoning effort instead leaning towards last-$k$ matches (Supplementary Figure~\ref{fig:sup_llm_matched}B and Supplementary Table~\ref{tab:sup_llm_ruletypes}). Similarly to people, models find simpler, often incomplete rules that nonetheless capture meaningful aspects of the underlying task structure. These findings suggest that the latest generation of more capable LRMs make more systematic errors than was previously found~\citep{legrisHARCRobustEstimate2024,mitchellComparingHumansGPT42023}.

Furthermore, we find that the distributions of error types show a similar correlation with performance. As in humans, \emph{attribute misselection} (see \S\ref{sec:results-taxonomy} for the taxonomy) is the most frequent error type across models, and its prevalence increases with performance. It accounts for $63.66\%$ ($94\%$ HDI [$60.63\%, 67.16\%$]) of Qwen3.5-27B errors and $61.58\%$ ($94\%$ HDI [$58.70\%, 64.62\%$]) of GPT-OSS-120B errors at high reasoning effort, compared with $49.11\%$ ($94\%$ HDI [$46.79\%, 51.56\%$]) and $42.63\%$ ($94\%$ HDI [$39.60\%, 45.38\%$]) at medium and low effort, respectively. Lower-performing models instead produce a larger proportion of \emph{unclassifiable} errors (Supplementary Figure~\ref{fig:sup_llm_matched}A and Supplementary Table~\ref{tab:sup_llm_categories}), suggesting less systematicity.

Finally, we investigate whether meaningful structural differences can differentiate human and LRM errors based on their synthesized programs. We trained classifiers to predict whether an error was made by a human or an LRM based on simple, linearized program-level features representing primitive count changes (see Figure~\ref{fig:llm-error-analysis}D and \S\ref{sec:methods-source-prediction} for a description of these features). We fit logistic L2 regression, explainable boosting machine (EBM) and gradient boosting classifiers. To avoid the confound of between-task program differences, we report ROC-AUC among each shared task's human and LRM errors and average across tasks. All three classifiers discriminate human from pooled LRM errors above chance and perform comparably (within-task ROC-AUC $=0.65$--$0.69$; Figure~\ref{fig:llm-error-analysis}E). Classifiers trained separately on humans versus each LRM configuration reliably discriminate LRM configurations which are less accurate than people (GPT-OSS-120B medium and low: $0.68$--$0.77$ across classifiers, all 95\% CI lower bounds $\geq 0.59$). However, discrimination is weaker and not reliably above chance for configurations which are more accurate than people (Qwen3.5-27B: $0.58$--$0.61$; GPT-OSS-120B high: $0.57$--$0.63$; Supplementary Table~\ref{tab:sup_llm_source_prediction_by_config}). Consistent with coarse feature representations, inspection of fitted coefficients and EBM shape functions indicates that this signal is distributed across many weak cues rather than a few diagnostic operations. We find that edits that remove higher-order primitives (e.g., \texttt{fork}, \texttt{chain}, \texttt{mapply}) or insert arithmetic (e.g., \texttt{double}, \texttt{increment}) are relatively important features for predicting LRM errors. Arithmetic insertions appear in about 5\% of human errors versus 10--16\% for each LRM configuration. Conversely, skipping a single drawing or recoloring step (e.g., \texttt{fill}, \texttt{paint}) is more characteristic of human errors (about 12\% vs.\ 2--9\%).
More generally, we find that people and LRMs tend to fail on different tasks: per-task success rates agree only weakly between people and each LRM configuration (Spearman $\rho = 0.30-0.34$, 95\% CIs within $[0.20, 0.44]$, across the 400 ARC-AGI-1 training tasks), substantially less than the LRM configurations which tend to agree with one another ($\rho = 0.49-0.82$). Although human and machine errors share broad patterns in simplicity bias, rule-type distributions, and error categories, measurable differences remain in the specific primitive operations that are used to reproduce their respective errors and problem-level difficulty profiles. This suggests that potentially different inductive biases support rule-search in people versus reasoning language models. Together, these results illustrate how BCSS can support investigating similarities and differences in human and machine reasoning through a shared symbolic representational space. 

\section{Discussion}\label{sec:discussion}

Symbolic models are psychologically motivated, interpretable, and directly comparable, but are difficult to scale beyond tasks with approximately enumerable hypothesis spaces. We address this scaling challenge by proposing BCSS. BCSS uses an LLM as a proposal distribution over program edits to recover, for individual responses, the rules people and machines may have applied in an open-ended domain. Three regularities emerge from the human data. Programs explaining errors are systematically shorter than the target rule, beyond the shortening attributable to the procedure itself. Many modeled errors appear to be explained by mental rules that fit some but not all of the training examples. And the most common error type is attribute misselection: computing a target parameter from the wrong feature. Together, these patterns are consistent with satisficing \citep{simonRationalChoiceStructure1956}: settling on a rule that fits enough of the evidence rather than searching for one that fits all of it.

Large reasoning models show the same broad signatures as people, graded by resources. In GPT-OSS-120B, lowering reasoning effort produces errors explained by simpler programs and fewer
partial fits to the evidence, paralleling human results. In combination with other research demonstrating a positive correlation between reasoning token allocation and thinking time from human responses in the H-ARC dataset \citep{devardaCostThinkingSimilar2025}, these results suggest that the patterns observed here reflect limits on search in both humans and LRMs. Yet, people and models fail on largely different tasks, and their error programs remain partially discriminable, pointing to differing inductive biases. Together, our results hint at a resource-rational characterization of reasoning in open-ended domains \citep{liederResourcerationalAnalysisUnderstanding2020}, though establishing this formally would require explicit modeling of the costs people and machines trade off \citep{kendallComplexityFormingMental2024}.

Modeling incorrect responses was central to the results we report here. Correct answers in isolation reveal little about how a solution was reached, whereas understanding systematic errors can enhance our explanations of underlying mechanisms \citep{tverskyJudgmentUncertaintyHeuristics1974,marcusOverregularizationLanguageAcquisition1992a,khemlaniTheoriesSyllogismMetaanalysis2012}. The order asymmetry among partial fits illustrates this. Exact Bayesian inference is order-invariant, whereas sequential sampling algorithms predict a dependence on presentation order
\citep{sanbornRationalApproximationsRational2010}, and because all examples remain visible throughout solving, degraded encoding is an unlikely cause of the observed patterns. An alternative is that task designers tend to place simpler examples first, so that a rule ignoring a later exception fits the early examples by construction. However, whether people form hypotheses from a limited subset of the examples, or form them more broadly but only verify them sequentially against part of the evidence remains an open question for future work.

Like much prior work in cognitive science \citep{chomskyAspectsTheorySyntax1965,marrVisionComputationalInvestigation1982,andersonAdaptiveCharacterThought1990}, BCSS treats inferences as deviations from a normative model. Here the normative standard takes the form of a reference program expressed in a DSL, with a locality prior keeping error programs within a few line edits of it. This strategy for understanding incorrect inferences has several precedents, for example in accounts of arithmetic errors as ``bugs'' in a correct procedure
\citep{brownDiagnosticModelsProcedural1978,vanlehnMindBugsOrigins1990}, and approaches that attribute rules to individuals from a pre-specified candidate set of strategies, distractors, or grammar-generated hypotheses \citep{sieglerThreeAspectsCognitive1976,kundaErrorPatternsRavens2016,nosofskyRuleplusexceptionModelClassification1994,goodman2008rational}. Conditioning an LLM proposal
distribution on the observed response removes the need to enumerate hand-design that set. However, one caveat is that each reference program is only one of many equivalent encodings of the
target rule, so edit distance and relative length are defined with respect to that encoding. The \verb|training| references were written independently of this work
\citep{hodelAddressingAbstractionReasoning2024}, and the main effects replicate on the \verb|evaluation| split, whose references were obtained by an entirely different
procedure (\S\ref{sec:synthetic-solvers}).

BCSS has further limitations. First, it captures about half of non-trivial errors, and because edits replace lines rather than add them, errors that elaborate on the
target rule are likely underrepresented in the analyzed set. It is also possible that these responses require more edits, or simply reflect more idiosyncratic behavior not governed by coherent rules. Second, the DSL was designed for solving ARC rather than for cognitive modeling. Although many of the operations in it are cognitively plausible (e.g., geometric transformations, object groupings, and other spatial predicates), some primitives (e.g.,
type coercion) have less clear cognitive counterparts. Consequently, the correspondence between cognitive complexity and DSL program complexity is unlikely to be one-to-one. Library learning, potentially guided by language
\citep{ellisDreamCoderBootstrappingInductive2021,wongLeveragingLanguageLearn2021,acquavivaCommunicatingNaturalPrograms2022} and an objective that encodes agreement with human thinking time, could instead automatically infer them. Relatedly, several
distinct programs often reproduce the same response, and we do not yet weight them, for instance by complexity or cognitive cost \citep{kendallComplexityFormingMental2024}. Third, programs are proposed and classified by LLMs from the same family, whose editing distributions may shape the categories we observe. Finally, BCSS identifies which rule an agent may have executed. Although we recover aspects of process-level signatures such as order-effects, our pipeline does not explicitly model how an agent arrived at that rule. Addressing process-level considerations will require integrating dynamic behavioral metrics such as response time, eye-tracking as well as carrying out controlled experimental conditions to further illuminate how human reasoners search and revise candidate rules. For LRMs, reasoning traces potentially offer a complementary and analogous window.

BCSS builds on multiple strands of research from AI and cognitive science. Recent advances have led to impressive developments in science automation \citep{novikovAlphaEvolveCodingAgent,fawziDiscoveringFasterMatrix2022,jansenCodeScientistEndtoEndSemiAutomated2025,ghareebMultiagentSystemAutomating2026,romera-paredesMathematicalDiscoveriesProgram2024}, with particular developments of this line of research to engineer systems for automated cognitive model discovery \citep{petersonUsingLargescaleExperiments2021,rmusGeneratingComputationalCognitive2025,castroDiscoveringSymbolicCognitive2025,jagadishClosingLoopDiscover2026,prystawskiAutopsychAutomatingScience2026,wangDiscoveringDifferencesStrategic2026}. Here, we develop similar tools to scale longstanding methods of psychological inquiry which have often faced a scalability bottleneck \citep{watermanProtocolAnalysisTask1971,ericssonProtocolAnalysisVerbal1993,wurgaftScalingThinkaloudMethod2025}. Inspired by seminal work such as Newell \& Simon's \cite{newell1972human} early attempts to develop theories of human problem-solving by analyzing detailed, individual-level behavioral protocols, we instantiate a similar but automated approach using AI tools. By using neural models as proposal distributions over discrete program spaces, BCSS makes symbolic, individual-level accounts
of behavior feasible in open-ended domains. Further, our approach demonstrates that cognitive footprints can be decoded from the structure of synthesized programs that explain behavior in different systems, offering a common ground for comparing how people and black-box AI systems solve complex problems. We see exciting prospects for the application of BCSS to other domains, shedding light on both human and machine cognition.

\section{Methods}\label{sec:methods}

\subsection{Data and filtering}\label{sec:methods-data}

We use the H-ARC dataset to understand how people reason on abstract rule-induction problems \citep{legrisComprehensiveBehavioralDataset2025, chollet2019measureintelligence}. The dataset consists of output grid responses across 800 distinct ARC-AGI-1 few-shot visual rule inference problems from over 1700 online participants. For the scope of the human results reported here, we focus our modeling efforts on a total of 723 ARC-AGI-1 tasks (362 tasks from the \verb|training| split; 361 from the \verb|evaluation| split), yielding a total of 6261 unique incorrect output grids across 1387 participants and 7402 attempts.

We pre-processed the H-ARC dataset by identifying all near-misses and ``trivial'' errors, corresponding to $15.6\%$ (1,222/7,809) of the full set of incorrect unique grids, and filtered them out under a noise tolerance criterion. Qualitative analyses revealed that for larger grids, ``errors'' could often be plausibly attributed to procedural mistakes (i.e., mis-clicks; Supplementary Figure~\ref{fig:sup_tolerance_examples}).  Given our modeling approach and goals, capturing this kind of noise within programs would not be informative about rule inference since we aim to understand conceptual mistakes. As such, we apply a noise tolerance criterion such that for any ``incorrect'' response grid of ten or more cells, we filter it out as a ``near-miss'' if it matches the correct test output grid with up to two mismatched cells; of the $353$ unique near-miss grids we found, $51.3\%$ differed from the correct output by a single cell and $47.9\%$ by exactly two (Supplementary Figure~\ref{fig:sup_coverage}, Supplementary Table~\ref{tab:sup_filtering}). Additionally, participants sometimes copied the test input grid ($7.9\%$ of unique responses) or training inputs and outputs ($3.5\%$ of unique responses). For test-input copies we applied the same tolerance criterion---for instance, if a participant produced a $5\times2$ output identical to the test input grid modulo one or two incorrect cells, we filtered it out as an identity error, whereas copies of a training example were removed only on an exact match. For grids smaller than ten cells, we apply a strict filter by only removing grids that are exact copies of the test input or training examples. Finally, we remove all ``empty'' 3$\times$3 grids ($1.9\%$ of
unique responses) because this grid state corresponds to the initial state of the test output grid in our task interface (i.e., a blank response).

\subsection{Domain-specific language and reference programs}\label{sec:methods-dsl}

We model rules as programs in a domain-specific language (DSL) for ARC. To do so, we leverage an expressive open-source DSL which was released alongside 400 human-written reference programs implementing the correct rule for each task \citep{hodelAddressingAbstractionReasoning2024}. The DSL consists of 160 primitive functions spanning object detection and manipulation (e.g., \verb|objects|, \verb|ofcolor|, \verb|recolor|, etc), grid transformations (e.g., \verb|rot90|, \verb|fill|, etc), spatial queries (e.g., \verb|height|, \verb|adjacent|, \verb|manhattan|, etc), selection and ranking (e.g., \verb|argmax|, \verb|mostcommon|, etc), and higher-order compositional and programmatic operations (e.g., \verb|compose|, \verb|fork|, \verb|branch|, etc).

The DSL is typed over a small set of object-centric types built from Python immutable types: grids as nested tuples of integers, objects as nested frozen sets of colored cells, etc. Additionally, the DSL contains a small set of primitive constants: a ten color palette, boolean values, and directional constants.

Programs in this DSL are written in A-normal form (ANF; \citep{flanaganEssenceCompilingContinuations1993}), a canonical form in which every non-trivial sub-expression is named via direct assignment, and function arguments must be atoms (i.e., variables or constants). The programs also respect static single assignment, where every variable is named once and only once. Although there are many possible expressions of correct programs for a given task rule, we focus on using the 400 human-written ones available from the open-source DSL for the \verb|training| split. For the remaining 400 tasks from the \verb|evaluation| split, we use the shortest synthesized solver for tasks where we successfully synthesized at least one solver.

\subsubsection{Synthesizing additional reference programs for full coverage}\label{sec:synthetic-solvers}

As described above, the DSL we utilize in this work was released with 400 human-written solvers for half of the publicly available tasks from ARC-AGI-1, excluding the 400 tasks in the \verb|evaluation| split. To obtain full coverage of H-ARC incorrect attempts, we prompted an LLM to generate synthetic solvers for the remaining 400 tasks. We successfully generated solvers for 361 tasks of the 392 tasks from the \verb|evaluation| split of ARC-AGI-1 that have incorrect responses. To do so, we simply prompted \verb|Gemini Pro 3.1| (Supplementary Figure~\ref{fig:sup_prompt_solver}) using retrieval-augmented generation \citep{lewisRetrievalAugmentedGenerationKnowledgeIntensive2020} with five in-context examples of other tasks and corresponding human-written solvers from the \verb|training| split. For each queried \verb|evaluation| task, we sampled five independent responses, and selected the shortest program from the set of valid and correct DSL programs for each task. Incorrect responses on the remaining 31 evaluation tasks (326 of the 3,892 non-trivial unique errors in that split) could not be modeled and are excluded, which leaves the 6,261 unique errors (2,695 training, 3,566 evaluation) analyzed here.

\subsection{Symbolic model synthesis}\label{sec:methods-pipeline}

\textbf{Behavior-Conditioned Symbolic Synthesis.} BCSS involves automating inference over behavioral outputs to produce programs representing the rules that an agent may have used to generate its responses. To so do, we leverage an LLM by prompting it to synthesize a program representation of each incorrect attempt. We sample from the LLM structured outputs of line edits restricted to a JSON schema, and its responses on task grids, a complete DSL reference, the reference program for the in-context task, and an incorrect output grid produced by a human participant or an LLM (Supplementary Figure~\ref{fig:sup_prompt_synthesis}). The LLM is instructed to generate a restricted set of edits to the in-context reference program such that the resulting program would produce the observed response when executed on the test input grid. The edits consist of in-place edits to referenced lines of code from the reference program. The LLM is instructed to generate syntactically valid, ANF lines of code written exclusively using the DSL. For each unique incorrect response, we collect 5 independent samples with edit budgets of 2, 4 and 8 lines, yielding a total of 15 samples per error. For human responses, we obtained a total of 93,855 program samples (40,365 for the \verb|training| set split responses; 53,490 for the \verb|evaluation| set split responses) across 723 tasks. Four training-split errors (five attempts) were missing from the synthesis run and thus were not sampled; they are retained in the error counts reported here.  Samples were generated using \verb|Gemini 3 Pro| for the \verb|training| split and \verb|Gemini 3.5 Flash| for the \verb|evaluation| split, with temperature set to 1.0, reasoning budget set to high, and default parameters otherwise.

For each sampled structured output of edits, we programmatically replace each tagged line with its corresponding edited line, and verify that the resulting programs respect ANF syntactic constraints and exclusively use DSL operators. Any edits that violate the instructions are rejected as invalid and discarded. Often, editing one or multiple lines has the side-effect of disconnecting variables in the reference program from the computation graph processing the input into the output grid. Thus, each program is rewritten using a custom parser which removes these pieces of code after applying the edits. We then execute each program using the test input grid for the corresponding task and verify its output against the target error grid. We only accept a program as an explanation for a given error if it produces an output that is equivalent to the closest match from the pool of generated programs under the same noise tolerance criterion as for filtering errors (\S\ref{sec:methods-data} for more details). Concretely, we apply a noise tolerance criterion such that for any synthesized program, if its output grid has ten or more cells, we consider it a valid model of the observed response if it matches the target response grid with up to two mismatched cells. For example, for a given target output grid, if the best program in the pool of 15 samples produces a correctly sized output that is at least 10 cells large, but differs by a single cell from the target, then any other program from that pool must produce outputs that are equally accurate to be accepted. After filtering, we obtained a total of 8,246 unique valid programs explaining human errors (3,690 for the \verb|training| set split responses; 4,556 for the \verb|evaluation| set split responses; see \S\ref{sec:methods-llms} for machine errors). Across modeled errors, we computed the per-error average minimum edits: the minimum number of edits averaged across errors. This serves as an upper bound on the shortest path in line-edit space from the reference program for explaining a given response.

\textbf{Exhaustive depth-1 program enumeration.} To characterize the space of programs adjacent to human-written ARC solutions, we exhaustively enumerated all single-edit mutations of the 400 \verb|training| split reference programs. Because the programs are written in ANF and respect static single assignment, the set of syntactically well-formed edits is finite and enumerable. We adapted code from previous work \citep{butt2024codeit} where a stochastic search algorithm over DSL program mutations in the ARC setting was proposed as a form of data augmentation to train deep learning models for program induction.

Each program is parsed into an abstract syntax tree and a concrete type is inferred for every variable by propagating the types of the task’s first training input through the DSL’s primitive signatures. For every assignment statement in the program, we then enumerated four kinds of type-correct edits:

\begin{itemize}
    \item Argument replacement: substituting one argument of the call for any other in-scope variable or primitive constant of the same type
    \item Function replacement: substituting the called primitive for any primitive with a compatible output type
    \item Right-hand-side replacement: substituting the entire expression for any type-correct application of a DSL primitive to in-scope variables and constants
    \item Argument replacement with variable introduction: inserting a new assignment immediately before a given statement and passing its result as the argument, with downstream variables renumbered accordingly
\end{itemize}

This procedure yielded a total of 16,661,032 candidate programs. We then executed each candidate on all input grids of its parent task. To do so, we wrote a custom line-by-line interpreter with a 0.25s per-statement timeout and a 1000-cell bound on grid and intermediate-state size, discarding executions that raised, diverged, or exceeded the size bound. Only candidates producing a valid output grid on both the training and test inputs were retained. This left 9,422,295 executable programs (56.6\% of those enumerated). Enumeration and execution were parallelized across 52 CPU cores and completed in approximately 16 hours on the NYU high-performance computing cluster. Programs were determined to match a human error under the same noise tolerance criterion as LLM-generated programs.

\subsection{Error classification}\label{sec:methods-classification}

To characterize the kinds of inferential shortcuts that people and machines utilize to solve ARC tasks, we developed a taxonomy of error types based on comparisons between programs that explain human and machine responses, and the reference programs from which they were adapted (see Table~\ref{tab:error-taxonomy}). Although some errors only require single-line edits, or even simpler node-level edits such as swapping an argument or a primitive function for another, the majority of synthesized programs that reproduce human responses required multi-line edits. Such program differences are difficult to programmatically interpret since their consequences can span different parts of the computation graph, and as a result they cannot be easily attributed to a particular high-level cognitive category without semantic interpretation or the contextual information from the training examples. Thus, we developed a precise taxonomy of seven error categories which we believed would capture meaningful cognitive constructs, and prompted an LLM (Supplementary Figure~\ref{fig:sup_prompt_classification}) to categorize each program from the final filtered collection of programs that match human or machine responses.

For each program, we prompt an LLM to return a structured JSON schema output containing one of eight possible labels: seven cognitive categories and an ``unclassifiable'' catch-all category. In addition to definitions for each category, we include the following elements in the model's prompt: the reference program for the given task, the task training input and output grids, the synthesized program, and the synthesized program execution trace on the training and test input grids. This final portion explicitates how the program behaves on the training examples, as well as what test output grid it produces (i.e., a human or machine response).  Concretely, the classifier must assign a label to structural differences between the reference program and the error program while taking into account the task data. For instance, a sub-tree that computes height relative to color count might instead be collapsed to a constant integer value in the error program, which coincides with some training examples. We sampled 5 independent classifications per program, yielding 18,450 classifications for the 3,690 programs that explain human errors on the \verb|training| split, and 41,655 classifications for the 8,331 programs that explain machine errors (4,115 for Qwen3.5-27B; 4,720, 13,800, and 19,020 for GPT-OSS-120B at high, medium, and low reasoning effort, respectively). All samples were generated using \verb|Gemini 3.1 Pro|, with temperature set to 1.0, reasoning budget set to high, and default parameters otherwise.

\subsection{Pipeline validation}\label{sec:validation}

To assess the validity and sources of bias of BCSS, we manually constructed a validation set of synthetic errors by mechanically producing targeted, simple edits on the full set of human-written and synthetic reference programs. For each of the eight categories, we defined a function that implements a limited set of mutation operators, applies them to a given input program, and returns new, modified programs that fall under the syntactic constraints of the corresponding error category. In total, this yielded $2{,}375$ synthetic errors across the 8 categories. We then manually verified a small random subset of errors for each category (444 in total) to obtain a target of $\sim\!20$ errors per category, balanced between the human-written solvers of the \verb|training| split and the LLM-synthesized solvers of the \verb|evaluation| split. Manual verification is required to ensure semantic interpretation of errors conditioned on task grids. In total, we obtained 148 synthetic errors which we use for validation, with attribute misselection containing 19 programs, and sequencing only 9. We refer to the errors generated and verified using the procedure just described as validation errors/programs. Validation errors refer to the grids obtained by executing the validation programs.

We passed the validation errors through BCSS end-to-end to evaluate how accurately the predicted categories matched the gold-labeled ones. Overall, this resulted in an end-to-end category agreement of $64.9\%$, $75.0\%$, and $76.4\%$ at edit budgets of $k \in \{1, 2, 3\}$, respectively. We evaluated these budgets because in this synthetic setting, program edit distances are known by construction---$124$ of the $148$ errors are single-line edits, $141$ are reachable within two edits, and all $148$ within three---though in practice the true edit distance is unknown. To understand what governs this performance and why accuracy plateaus around $76\%$, we decompose the pipeline into three potential sources of loss: classification error (\emph{labeling}), search failure (\emph{recovery}), and semantic shift under alternative programs (\emph{aliasing}).

We first isolated labeling by running the classifier directly on the validation programs, which establishes an accuracy ceiling of $85.1\%$. Classifications across five runs per program were highly consistent at $\sim\!95\%$ agreement, and we used modal labels as the unit of analysis. Second, we evaluated whether search failure accounts for the drop from this ceiling. We found that given an adequate edit budget, it did not account for drop in accuracy. Increasing $k$ from 1 to 3 raised recovery from $84.5\%$ to $99.3\%$ ($95.3\%$ at $k = 2$), successfully finding a candidate program for all but one validation error grid. Instead, the remaining gap is driven almost entirely by what we call ``program aliasing'': functionally equivalent programs that are written differently. For instance, consider a task in which a set of isolated cells must be expanded into a $3\times3$ square of a new color. An error in which the central cell keeps its original color can be produced by restricting the recoloring to background cells surrounding the center cell (an injected \textit{scope restriction}), but also by omitting the step that fills the square's interior so that only the ring around the cell is recolored, a program that the classifier instead labels as a \textit{partial program}. Across all budgets, BCSS recovered the exact injected program in roughly two-thirds of cases ($62\%\text{--}65\%$). When search found an alternative, aliased program instead, the classifier still identified the injected category in $57.3\%$ of cases (remaining essentially flat at $52.6\%\text{--}59.4\%$ across budgets). Thus, program aliasing rather than search failure or classifier noise, accounts for the primary drop between the $85.1\%$ ceiling and end-to-end performance.

\subsection{Evaluating machine errors}\label{sec:methods-llms}

As described in \S\ref{sec:llm-errors}, we evaluate two reasoning-tuned LLMs on the \verb|training| split of ARC-AGI-1 to study their error behavior: Qwen3.5-27B~\citep{qwen3.5} in thinking mode and GPT-OSS-120B~\citep{openai2025gptoss120bgptoss20bmodel} at high, medium, and low reasoning effort. We evaluate each model configuration across ten random seeds using a standard evaluation prompt (see Supplementary Figure~\ref{fig:sup_prompt_llm_eval}), on all $416$ test inputs of the $400$ tasks ($4{,}160$ responses per configuration). H-ARC participants attempted only the first test input of the $14$ tasks that have two so we only consider responses to that attempt in our analyses. For Qwen3.5-27B, we follow the recommended sampling settings for general tasks in thinking mode\footnote{\url{https://huggingface.co/Qwen/Qwen3.5-27B}} with a temperature of $1.0$, a top-$p$ value of $0.95$, a top-$k$ value of $20$, a minimum-$p$ value of $0.0$, a presence penalty of $1.5$, and a repetition penalty of $1.0$. For GPT-OSS-120B, we follow GPT-OSS's official vLLM example\footnote{\url{https://developers.openai.com/cookbook/articles/gpt-oss/run-vllm}} and use a temperature of $1.0$, a top-$p$ value of $1.0$, disabled top-$k$ filtering, a minimum-$p$ value of $0.0$, a presence penalty of $0.0$, and a repetition penalty of $1.0$ across all three reasoning-effort settings. We set a maximum output budget of $65{,}536$ generated tokens, including reasoning and the final answer, for both models. For GPT-OSS-120B, we control reasoning behavior through the reasoning-effort setting without imposing a separate reasoning-token budget. For Qwen3.5-27B, we additionally impose a reasoning-token budget of $49{,}152$ tokens to limit time and resources spent on inference. If Qwen3.5-27B reaches this budget before ending its reasoning, the end-of-thinking marker \texttt{</think>} is appended to its current output, terminating the reasoning block and allowing final-answer generation within the remaining output budget. We observe that only $4.35\%$ ($181 / 4160$) of responses reach the model's maximum reasoning budget.

A response is counted as correct only if generation finishes within the maximum token budget, and the extracted grid exactly matches the target in both dimensions and cell values. We parse predicted grids by selecting the last valid group of bracketed grid rows, accepting spaces or commas between entries, and falling back to parsing Python/JSON-style nested lists. A valid grid must be nonempty, rectangular, and contain integer values from 0 to 9. Completed responses from which no valid grid can be extracted are classified as parsing errors. Both unfinished responses and completed responses with no valid grid remain included in the overall accuracy calculations but are excluded from subsequent analyses (see Supplementary Table~\ref{tab:sup_llm_yield}). We find that Qwen3.5-27B generates $806$ incorrect responses, of which $689$ contain valid but incorrect grid predictions. For GPT-OSS-120B, $742 / 1153$ incorrect responses at high reasoning effort contain valid but incorrect grid predictions, compared with $2002 / 2040$ at medium reasoning effort and $3059 / 3119$ at low reasoning effort. After removing incorrectly formatted grid outputs, we applied the additional filtering steps described in~\S\ref{sec:methods-data} (see Supplementary Table~\ref{tab:sup_llm_filtering} for the full per-configuration breakdown).

We pass each incorrect grid prediction to BCSS as described in~\S\ref{sec:methods-pipeline}. As with human responses, we collect $5$ independent program samples at edit budgets of $2$, $4$, and $8$ lines, yielding a total of $15$ samples per machine error. Unlike for humans, identical grids from different seeds were sampled separately, but programs were subsequently deduplicated and pooled per unique error. In total, we obtain $8,625$ program samples for errors from Qwen3.5-27B and $9,780$ samples for GPT-OSS-120B at high reasoning effort, $26,070$ at medium reasoning effort, and $37,365$ at low reasoning effort. At the time when we ran machine evaluations, \texttt{Gemini 3 Pro} was unfortunately not available anymore on the Gemini API. Thus, samples are generated using \texttt{Gemini 3.5 Flash}, with the temperature set to $1.0$, the reasoning budget set to high, and all other parameters left at their default values. All programs are subsequently verified and filtered as described in~\S\ref{sec:methods-pipeline}. After filtering, we obtain $823$ unique valid programs for Qwen3.5-27B's errors, $944$ programs describing GPT-OSS-120B's errors at high reasoning effort, $2,760$ at medium reasoning effort, and $3,804$ at low reasoning effort. For comparisons across models and humans, we restrict our analyses in~\S\ref{sec:llm-errors} to the $56$ tasks on which humans and every model configuration make at least one error, thereby avoiding task difficulty as a confounding factor. On these $56$ tasks, Qwen3.5-27B produces $112$ errors captured by $486$ programs, whereas GPT-OSS-120B produces $128$ errors described by $608$ programs at high reasoning effort, $221$ errors described by $1,008$ programs at medium reasoning effort, and $206$ errors described by $916$ programs at low reasoning effort. On the same tasks, humans produce $144$ first-attempt errors described by $450$ programs. All statistical models in~\S\ref{sec:methods-stats} are fit separately for each model configuration and for humans, using only the first-attempt errors and programs from these $56$ tasks unless stated otherwise (\S\ref{sec:simplicity-model}).

\subsection{Predicting human versus machine errors}\label{sec:methods-source-prediction}
We train classifiers on program features to predict whether an error was made by a human or a large reasoning model (LRM). As described in~\S\ref{sec:methods-pipeline}, DSL programs are written in A-normal form as sequences of primitive function calls. We describe each error program by how it differs from its reference program (Figure~\ref{fig:llm-error-analysis}D).

Primitive delta vectors $\mathbf{v}_{\Delta} \in \mathbb{Z}^{160}$ record the signed difference in call counts between the error-producing and reference programs, with one entry per DSL primitive. For example, if one \texttt{rotate180} call is replaced by \texttt{hmirror} and all other calls remain unchanged, the vector contains $-1$ for \texttt{rotate180}, $+1$ for \texttt{hmirror}, and zeros elsewhere.

We fit L2-regularized logistic regression, explainable boosting machines (EBMs), and histogram gradient boosting classifiers on the extracted features. Each candidate program explaining an observed error provides one training example. Source-balanced weights allocate half the total training weight to humans and half to LRMs, with the LRM share divided equally among the four LRM configurations. Within each source, the weight is divided equally among tasks, then among errors within each task, and finally among candidate programs within each error. Logistic regression features are scaled by training-fold standard deviations without centering; boosting models use raw values.

Classifiers are trained on errors from all tasks, which maximizes the number of examples available for training. Because classifiers are fit at the candidate-program level, we obtain error-level predictions by averaging probabilities across all candidate programs explaining a given error. We use nested cross-validation with five outer and five inner folds, keeping all errors and candidate programs from a task together at both levels to avoid information leakage. The outer and inner folds are defined over the 56 shared tasks on which humans and every LRM configuration make at least one modeled error (\S\ref{sec:methods-llms}) and the remaining 304 tasks are included as training data in every outer and inner fold. Thus, reported metrics are computed only on the out-of-fold predictions for the 56 shared tasks. In the outer loop, one fold of shared tasks is held out for testing and the remaining shared tasks are partitioned into the five inner folds for hyperparameter selection. We select hyperparameters to maximize mean within-task ROC-AUC across the five inner validation folds. The winning hyperparameter configuration is then used to refit the classifier on the full outer training split and generate out-of-fold predictions on the held-out outer test fold. We search the inverse L2 regularization strength $C \in \{0.01,0.1,1,10,100\}$ for logistic regression, $\{0,5\}$ pairwise interactions and $\{100,300\}$ boosting rounds for EBM, and $\{7,15\}$ maximum leaf nodes and $\{100,300\}$ iterations for histogram gradient boosting. Pre-processing and interaction selection use only data from the corresponding training fold.

We score predictions using within-task ROC-AUC, treating LRM errors as the positive class and averaging equally across tasks. We pool the outer test-fold predictions across all five outer splits, with 95\% percentile confidence intervals estimated from $10{,}000$ bootstrap resamples of whole tasks holding the out-of-fold predictions fixed. To examine whether discriminability differs across LRM configurations, we also fit each classifier to distinguish human errors from those of a single configuration, using the same features, hyperparameter grids, task folds, shared evaluation tasks, and bootstrap resamples, with the LRM half of the training weight assigned entirely to that configuration.

\subsection{Statistical analyses}\label{sec:methods-stats}

All analyses were performed using a Bayesian framework. Continuous outcomes were modeled as Gaussian linear mixed-effects regressions, using Bambi v0.17.2 for the main simplicity model and PyMC v5.28.5 directly for error category simplicity model; categorical outcomes were modeled as non-centered hierarchical categorical regressions directly in PyMC v5.28.5 with a softmax inverse link. All posteriors were approximated using the No-U-Turn Sampler (NUTS; \citep{hoffman2014no}) via the JAX/NumPyro backend \citep{phan2019composable}, with a target acceptance rate set to $0.9$. We ran inference using 4 chains, with warm-up samples specified for each model below. For all models, we use non-centered parameterizations of the random effects, which is standard practice for inference using NUTS. Convergence was assessed by the rank-normalized split-$\hat{R}$ statistic and bulk/tail effective sample size, using thresholds of $\hat{R} < 1.01$ and ESS $> 400$ per parameter; we additionally verified that no divergent transitions occurred. For every parameter of interest we report the posterior mean (or median for probabilities) and the 94\% highest-density interval (HDI).

\subsubsection{Simplicity bias}\label{sec:simplicity-model}

To determine whether error-producing programs were simpler than baseline programs synthesized for correct outputs under an identical synthesis procedure, we fit a hierarchical Bayesian linear model. Let $y_{ijk}$ denote the relative line-count change of the $k$-th synthesized program for unique error $j$ in task $i$, and let $x_{ijk} \in \{0,1\}$ indicate the treatment condition (where $x_{ijk} = 0$ for control and $x_{ijk} = 1$ for treatment). We compute the relative line-count change as $(L(p) - L(p^*)) / L(p^*)$, where $L(p)$ is the line count of the synthesized program, and $L(p^*)$ is the line count of the reference program from which it was adapted. We modeled the outcome as:

$$
  \begin{aligned}
    y_{ijk} \mid \mu_{ijk}, \sigma & \sim \mathcal{N}(\mu_{ijk}, \sigma^2),      \\
    \mu_{ijk}                      & = \beta_0 + \beta_1 x_{ijk} + u_i + v_{ij}, \\
    u_i                            & \sim \mathcal{N}(0, \tau_{\text{task}}^2),  \\
    v_{ij}                         & \sim \mathcal{N}(0, \tau_{\text{err}}^2)
  \end{aligned}
$$
We include task-level random intercepts $u_i$ to account for baseline variations in program length across ARC tasks, and nested per-error intercepts $v_{ij}$ to handle the fact that each unique error contributes multiple candidate programs that are not independent observations. Thus, the model answers whether synthesized programs that fit incorrect human or machine responses, tend to be more or less complex than reference programs subjected to an identical procedure modulo the conditioned response data. We dummy-coded the condition factor with the control baseline as the reference level, so that the slope $\beta_1$ is directly interpretable as the treatment-minus-control contrast. Analyses were restricted to tasks present in both conditions to ensure the random-effects structure was identified. We report the treatment and control mean relative line-count change ($\beta_0 + \beta_1$, $\beta_0$) and the treatment–control contrast ($\beta_1$) corresponding to additional shortening attributable to conditioning synthesis on incorrect responses.

Control programs were synthesized with the same model as the treatment programs they are compared against: \verb|Gemini 3 Pro| for human errors on the \verb|training| split, and \verb|Gemini 3.5 Flash| for human errors on the \verb|evaluation| split and for machine errors (\S\ref{sec:methods-pipeline}, \S\ref{sec:methods-llms}). For the comparison with machines (Figure~\ref{fig:llm-error-analysis}B), fits are restricted to the 55 of the 56 matched tasks (\S\ref{sec:methods-llms}) that have a control program. The four machine fits share a single control set of 150 programs, and the human fit uses 132 control programs.

To relate the simplicity bias to performance, we propagated posterior uncertainty in the treatment-minus-control contrasts into the correlation. For each of 20,000 posterior draws, we took the relative shortening $-\beta_1$ from each of the five fits, computed its Pearson correlation with the five accuracy rates, and fit an ordinary least-squares slope of shortening on accuracy. We report the posterior median and 94\% HDI of both quantities. Accuracy is treated as fixed, so these intervals reflect uncertainty in the shortening estimates only.

Finally, we placed weakly informative priors on all coefficients and variance components:
\begin{align*}
  \beta_0                                  & \sim \mathcal{N}(0, 1.5^2),         \\
  \beta_1                                  & \sim \mathcal{N}(0, 1.5^2),         \\
  \tau_{\text{task}}, \, \tau_{\text{err}} & \sim \mathrm{Half\text{-}Normal}(1)
\end{align*}
The residual standard deviation $\sigma$ was assigned \verb|Bambi|'s default data-scaled prior.

\subsubsection{Rule types and order effects}\label{sec:rule-type-model}

Given error-producing programs, we modeled how often such programs also produce the correct output grids for at least one training example but not all (i.e., partial-match programs), or all the training examples but not the test example (i.e., alternative rules). To do so, we fit a hierarchical categorical Bayesian model, and each program was run on all the training examples of its associated task and labeled according to the outputs it produced. Let $y_{i,j} \in \{1, \dots, K\}$ be the rule-type classification of program $j$ for unique error $i$, where $K= 3$ categories: ``alternative rule'' (set as reference, $k=1$), ``neither'' ($k=2$) and ``partial-match'' ($k=3$). We modeled the likelihood and linear predictors as:

$$
  \begin{aligned}
    y_{i,j} \mid \boldsymbol{p}_i & \sim \mathrm{Categorical}(\boldsymbol{p}_i), \\
    p_{i,k}                       & = \mathrm{softmax}(\boldsymbol{\eta}_i)_k,   \\
    \eta_{i,1}                    & = 0 \quad\text{(reference)},                 \\
    \eta_{i,k}                    & = \beta_{0,k} + u_{i,k} \quad k = 2,\dots,K, \\
    u_{i,k}                       & \sim \mathcal{N}(0, \tau_{k}^2)
  \end{aligned}
$$
where $\boldsymbol{p}_i = (p_{i,1},\dots,p_{i,K})$ is the error-specific probability vector on the $(K-1)$-simplex, shared by all programs $j$ belonging to error $i$. As in the previous model, we include per-error intercepts $u_{i, k}$ since there are often multiple programs per error, each with a possibly different behavior on the training examples. The ``alternative rule'' category is fixed at zero as the reference, since the softmax link is translation invariant. Finally, we placed weakly informative priors on the coefficient and variance components:

$$
  \begin{aligned}
    \beta_{0,k} & \sim \mathcal{N}(0,\ 2.5^2),        \\
    \tau_{k}    & \sim \mathrm{Half\text{-}Normal}(1)
  \end{aligned}
$$

With this model, we seek to understand how likely it is that an incorrect rule is modeled by a program that also satisfies some or all of the constraints imposed by the training examples human or machine reasoners observed. Accordingly, we define the attempt-weighted rule-type prevalence over the empirical error pool as a derived quantity of the model parameters:

$$
  P_k(\boldsymbol{\theta}) = \sum_{i=1}^{N} w_i \cdot
  \mathrm{softmax}\!\left([0,\ \eta_{i,2},\ \eta_{i,3}]\right)_k,
  \qquad w_i = \frac{c_i}{\sum_{i'} c_{i'}}
$$
where $\boldsymbol{\theta} = (\boldsymbol{\beta}_0, \boldsymbol{\tau}, \{\boldsymbol{u}_i\}_{i=1}^N)$, $\eta_{i,k}$ is defined as above, $c_i$ is the number of attempts that produced error $i$ (the number of participant attempts for human data; for model configurations, the number of sampled responses), and $N$ is the total number of unique errors. Evaluating $P_k$ at each posterior draw $\boldsymbol{\theta}^{(s)}$, $s = 1, \dots, S$, yields a Monte Carlo sample $\{P_k(\boldsymbol{\theta}^{(s)})\}_{s=1}^{S}$ from the posterior distribution of the prevalence, which we summarize by its median and 94\% HDI. Because each error contributes its realized random effect $\boldsymbol{u}_i$ weighted by $w_i$, $P_k$ corresponds to the expected rule-type prevalence one would observe when sampling incorrect attempts uniformly at random from the human or machine incorrect response distributions.
  
To characterize which output structure is reproduced by the partial-match rules, we fit a second instance of the same hierarchical categorical model, restricted to the partial-match programs, on ordering categories of the reproduced training pairs. For a partial-match program that reproduces $0 < k < n$ of the $n$ training outputs, we assign an ordering category based on the positions of the matched pairs: ``first-$k$'' when they form a block anchored at the first training example (similar to a primacy pattern), ``last-$k$'' when the block is anchored at the last example (similar to a recency pattern), ``contiguous'' when the run is interior (length $\ge 2$, not including the first or last  example), and ``other'' for a single interior match or a scattered, non-contiguous set. Let $o_{i,j} \in \{1,\dots,K'\}$ be the ordering category of partial-match program $j$ for unique error $i$, where $K' = 4$ and ``contiguous'' is the reference ($k=1$). The likelihood, softmax link, per-error random effects $u_{i,k}$, and priors are exactly as in the rule-type model above; only the observation set (partial-match programs), the category set, and hence the number of categories change. We summarize the same attempt-weighted prevalence $P_k(\boldsymbol{\theta})$ by its posterior median and 94\% HDI, with the weights $w_i = c_i / \sum_{i'} c_{i'}$ renormalized over the errors that admit at least one partial-match program. The resulting prevalences are therefore conditional on an error being explained by a partial-match program, and describe, among such attempts, how the reproduced training examples are distributed by position. For both models, we report only the attempt-weighted prevalences $P_k$, expressed as percentages. 

\subsubsection{Error categories}\label{sec:error-class-model}

Similarly to the rule-type model described above, we modeled the prevalence of the different error types among the programs that reproduce incorrect human or machine attempts on ARC tasks. For each program, we first obtain five classification samples using our LLM classifier (\S\ref{sec:methods-classification}). Given this pool of samples, each unique response may have between $5$ and $5M_{\max}$ label samples, where $M_{\max}$ is the largest number of unique error programs retained for a single error. Let $y_{ijs} \in \{1,\dots,K\}$ be the error category assigned by sample $s$ to program $j$ of unique error $i$, where $K = 8$ categories: ``attribute misselection'' (set as reference, $k=1$), ``boundary precision'' ($k=2$), ``sequencing'' ($k=3$), ``operator confusion'' ($k=4$), ``partial program'' ($k=5$), ``scope restriction'' ($k=6$), ``transformation misselection'' ($k=7$), and ``unclassifiable'' ($k=8$). We modeled,

$$
  \begin{aligned}
    y_{i,j,s} \mid \boldsymbol{p}_{ij} & \sim \mathrm{Categorical}(\boldsymbol{p}_{ij}),         \\
    p_{ij,k}                           & = \mathrm{softmax}(\boldsymbol{\eta}_{ij})_k,           \\
    \eta_{ij,1}                        & = 0 \quad\text{(reference)},                            \\
    \eta_{ij,k}                        & = \beta_{0,k} + u_{i,k} + v_{ij,k} \quad k = 2,\dots,K, \\
    u_{i,k}                            & \sim \mathcal{N}(0, \tau_{k}^2),                        \\
    v_{ij,k}                           & \sim \mathcal{N}(0, \omega_{k}^2)
  \end{aligned}
$$

The difference with the rule-type model is the additional program-level random effect $v_{ij,k}$, nested within the per-error effect $u_{i,k}$, and the number of categories. We include both effects because each unique error may be explained by several distinct programs, and each program is in turn labeled multiple times by the classifier. The per-error intercepts $u_{i,k}$ absorb systematic differences between errors, and the program-level intercepts $v_{ij,k}$ absorb variation between the programs generated for the same error together with the labeling noise across the replicated samples. As before, ``attribute misselection'' is fixed at zero as the reference, since the softmax link is translation invariant. We placed the same weakly informative priors on the coefficient and both variance components:

$$
  \begin{aligned}
    \beta_{0,k}          & \sim \mathcal{N}(0,\ 2.5^2),        \\
    \tau_{k}, \omega_{k} & \sim \mathrm{Half\text{-}Normal}(1)
  \end{aligned}
$$

As with the rule-type model, we define the attempt-weighted error-category
prevalence over the empirical error pool as a derived quantity of the model
parameters:

$$
  P_k(\boldsymbol{\theta}) = \sum_{i=1}^{N} w_i \cdot
  \mathrm{softmax}\!\left([0,\ \tilde{\eta}_{i,2}, \dots, \tilde{\eta}_{i,K}]\right)_k,
  \qquad w_i = \frac{c_i}{\sum_{i'} c_{i'}}
$$

where $\boldsymbol{\theta} = (\boldsymbol{\beta}_0, \boldsymbol{\tau},
  \boldsymbol{\omega}, \{\boldsymbol{u}_i\}_{i=1}^N, \{\boldsymbol{v}_{ij}\})$,
$\tilde{\eta}_{i,k} = \beta_{0,k} + u_{i,k}$ is the error-level linear predictor evaluated with the program-level effects held at zero ($v_{ij,k} = 0$), $c_i$ is the number of attempts that produced error $i$ (the number of participant attempts for human data; for model configurations, the number of sampled responses), and $N$ is the total number of unique errors. Evaluating $P_k$ at each posterior draw $\boldsymbol{\theta}^{(s)}$, $s = 1, \dots, S$, yields a Monte Carlo sample
$\{P_k(\boldsymbol{\theta}^{(s)})\}_{s=1}^{S}$ from the posterior distribution of the prevalence, which we summarize by its median and 94\% HDI. Holding $v_{ij,k} = 0$ evaluates each error at its typical program; the prevalences therefore describe the error category one would expect for a typical program explaining an incorrect attempt sampled uniformly at random from the human or machine incorrect response distributions. Similarly to rule types, we report the attempt-weighted category prevalences $P_k$.

\subsubsection{Relative length change and error categories}\label{sec:error-cat-diff-model}

Having established that error programs are simpler than the reference programs they are adapted from (\S\ref{sec:simplicity-model}), we next asked how this relative shortening varies with reference program length, task difficulty, and error type. We fit a Bayesian linear regression predicting the relative line-count change of each error program, with reference program line count and empirical task success rate as continuous covariates and the error category as an additional term. Since we only ran error classification on programs fitting incorrect responses to tasks from the \verb|training| split, this model is only run on that subset of programs. Let $y_{ijr}$ denote the relative line-count change (defined as in \S\ref{sec:simplicity-model}) of the $r$-th error program for unique error $j$ in task $i$, let $s_i$ be the H-ARC empirical success rate of task $i$, and let $\ell_i = L(p^*_i)$ be the line count of the task's reference program. Both covariates were mean-centered, with $\bar{s}$ and $\bar{\ell}$ denoting their sample means. We modeled the outcome as:
$$
  \begin{aligned}
    y_{ijr} \mid \mu_{ijr}, \sigma & \sim \mathcal{N}(\mu_{ijr}, \sigma^2),   \\
    \mu_{ijr}                      & = \beta_0 + \beta_1 (s_i - \bar{s})
    + \beta_2 (\ell_i - \bar{\ell})
    + \boldsymbol{\theta}_{ijr}^{\top} \boldsymbol{\gamma}
    + u_{ij},                                                                 \\
    u_{ij}                         & \sim \mathcal{N}(0, \tau_{\text{err}}^2)
  \end{aligned}
$$

As in the previous models, we include per-error random intercepts $u_{ij}$ to account for the several candidate programs that may explain a single unique error. Centering the covariates makes $\beta_0$ the expected relative length change at the mean success rate and mean reference length. The slopes $\beta_1$ and $\beta_2$ capture how relative shortening covaries with task difficulty and reference program complexity, respectively.

The vector $\boldsymbol{\gamma} = (\gamma_1, \dots, \gamma_K)$ is a per-category effect on relative length change for the $K = 8$ categories of \S\ref{sec:error-class-model}, and $\boldsymbol{\theta}_{ijr}$ is a fixed $K$-vector of category weights for program $r$. Instead of collapsing the classifier's five samples per program to a single majority-vote label, we carry that uncertainty into the covariate. Let $\boldsymbol{c}_{ijr}$ be the program's category count vector across its $n_{ijr}$ classification samples. We set $\boldsymbol{\theta}_{ijr}$ to the Dirichlet--multinomial posterior mean under a symmetric $\mathrm{Dirichlet}(\alpha)$ prior,

$$
  \theta_{ijr,m} = \frac{c_{ijr,m} + \alpha}{n_{ijr} + K\alpha}, \qquad \alpha = 0.1,
$$
so that the category effect enters the mean linearly as the probability-weighted average $\boldsymbol{\theta}_{ijr}^{\top}\boldsymbol{\gamma}$. We used a minimal smoothing constant ($\alpha = 0.1$) so that a program with unanimous classifications loads its weight almost entirely on its own category, keeping $\boldsymbol{\gamma}$ on the true per-category magnitude scale, with non-unanimous programs distributing weight across their sampled categories. Treating $\boldsymbol{\theta}_{ijr}$ as fixed data (rather than as a latent allocation informed by the length outcome) keeps the category term a plain linear-Gaussian covariate with unimodal geometry. Unlike the corresponding latent-mixture formulation which failed to converge, this model samples cleanly under NUTS.

The seven cognitive categories share a sum-to-zero prior, so that each of their effects is interpretable as a deviation from their common grand mean $\beta_0$. The ``unclassifiable'' category is held out of this constraint and given its own free effect. We set the model this way since this category collects short, degenerate programs whose shortening is several times larger than that of any cognitive category, and pooling it into the sum-to-zero prior would stretch that scale and distort the grand-mean anchor. We placed weakly informative priors on the coefficients and variance components:

$$
  \begin{aligned}
    \beta_0, \beta_1, \beta_2    & \sim \mathcal{N}(0,\ 1.5^2),                                                   \\
    (\gamma_1, \dots, \gamma_7)  & \sim \mathrm{ZeroSumNormal}(1.5), \quad \textstyle\sum_{m=1}^{7} \gamma_m = 0, \\
    \gamma_8                     & \sim \mathcal{N}(0,\ 1.5^2),                                                   \\
    \tau_{\text{err}}, \, \sigma & \sim \mathrm{Half\text{-}Normal}(1.5)
  \end{aligned}
$$

Because of the custom sum-to-zero structure on the category effects, this model was written and sampled directly in \verb|PyMC| rather than through \verb|Bambi|. We report the change in relative shortening per additional reference-program line ($\beta_2$) and per unit of task success rate ($\beta_1$) in percentage points. We also report each category’s predicted relative length change ($\beta_0 + \gamma_m$) and its deviation from the grand mean ($\gamma_m$), and the predicted relative length change at a given reference length ($\beta_0 + \beta_2(\ell - \bar{\ell})$).

\backmatter

\bmhead{Supplementary information}

This article includes Supplementary Information: Supplementary Notes~S1--S6, Supplementary Figures~S1--S8 and Supplementary Tables~S1--S10.

\bmhead{AI use statement}

We acknowledge the use of generative AI tools for minor language revisions, such as grammar
checks and reformulations. In addition, we used generative AI tools such as Claude Code and Codex to assist with
code implementation, review and testing. SL and PM reviewed all code generated by Claude Code and Codex and tested it for correctness. We take responsibility for the final content of this work, including text, claims, and artifacts produced with the aid of generative AI.

\bmhead{Funding}

SL was supported by a combination of NSF Award 2121101 to TMG, the NYU Training Program in Computational Neuroscience (NIH R90 DA060339), and the MacCracken Fellowship at NYU. 
PM was supported by the LMU–NYU Cooperation Program during his visit to NYU and Princeton University making this collaboration possible.
AKJ is supported by a Natural and Artificial Minds (NAM) Fellowship from the Scully Peretsman foundation at Princeton University.
We thank Google for generously supporting this research with a financial gift to BML by providing Google Cloud Credits to utilize Gemini models for research.

\bmhead{Data availability}
This study uses the publicly available H-ARC dataset~\citep{legrisComprehensiveBehavioralDataset2025}. All relevant data will be made publicly available with a permanent DOI upon publication.

\bmhead{Code availability}
Code for BCSS and all analyses is available at \url{https://github.com/Le-Gris/bcss}.

\bmhead{Author contribution} SL, AKJ, BML and TMG conceptualized the study. SL developed the methodology, implemented the software and performed the main analyses. JJ contributed to preliminary analyses. PM implemented the machine evaluations, and PM and SL analysed the machine results. SL created the figures. BML and TMG supervised the project, and AKJ provided guidance throughout. SL wrote the manuscript, with PM drafting the machine evaluation and analysis sections. AKJ, BML and TMG provided substantial feedback and edits, and all authors reviewed the manuscript.

  \renewcommand{\thesection}{S\arabic{section}}
  \renewcommand{\thesubsection}{S\arabic{section}.\arabic{subsection}}
  \setcounter{section}{0}
  \setcounter{figure}{0}
  \setcounter{table}{0}
  \renewcommand{\thefigure}{S\arabic{figure}}
  \renewcommand{\thetable}{S\arabic{table}}

  \renewcommand{\theHsection}{S\arabic{section}}
  \renewcommand{\theHsubsection}{S\arabic{section}.\arabic{subsection}}
  \renewcommand{\theHfigure}{S\arabic{figure}}
  \renewcommand{\theHtable}{S\arabic{table}}

  \section*{Supplementary Information}

  This supplement reports (i) data for understanding the noise-tolerance criterion
  (\S\ref{sup:tolerance}), (ii) per-split versions of every quantity reported
  as pooled results in the main text (\S\ref{sup:coverage}--\S\ref{sup:length}), (iii) per-model values for the human--machine comparison (\S\ref{sup:llm_alltasks}), and (iv) the prompts used at each stage of BCSS (\S\ref{sup:prompts}).

  \FloatBarrier

  \section{Filtering policy}\label{sup:tolerance}

  Our filtering policy (\S\ref{sec:methods-data}) treats an incorrect response of ten or more cells as a near-miss, and filters it out as approximately correct, when it differs from the correct test output in at most two cells. We apply the same criterion for synthesized programs that approximately reproduce an error (\S\ref{sec:methods-pipeline}). We provide some rationale for this two-cell constant in the pipeline.

\begin{figure}[p]
    \centering
    \includegraphics[width=\textwidth,height=0.85\textheight,keepaspectratio]{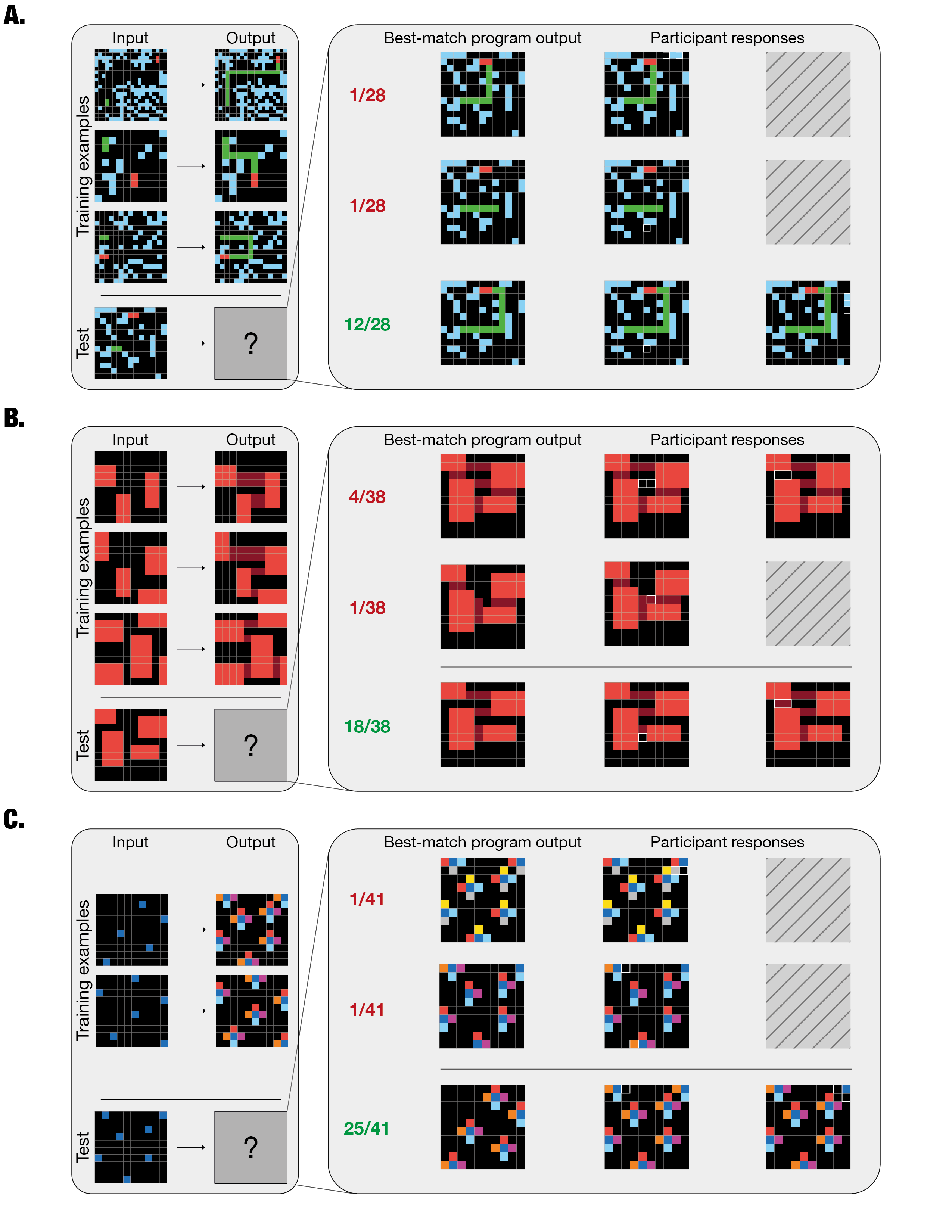}
    \caption{\textbf{Responses within the two-cell noise tolerance.}
      (\textbf{A}--\textbf{C}) Three representative ARC tasks. Left: the training examples and the test input shown to participants (the test output is hidden with a grey square and ``?''). Right: for three response grids per task, the output of the best-matching synthesized program (left column) beside the participant responses it is accepted as reproducing (remaining columns); hatched cells indicate that fewer than two distinct responses share that program. The fraction to the left of each row gives the number of participants producing that response out of all participants who attempted the task (red, incorrect; green, correct). Within each row the grids differ in at most two isolated cells (surrounded by white) against the best-match program output.}\label{fig:sup_tolerance_examples}
  \end{figure}

  Supplementary Figure~\ref{fig:sup_tolerance_examples} shows representative responses within the tolerance interval. In each case the participant's grid and the best-matching program output differ in one or two isolated cells against an otherwise identical structure.

\begin{figure}[p]
    \centering
    \includegraphics[width=\textwidth,keepaspectratio]{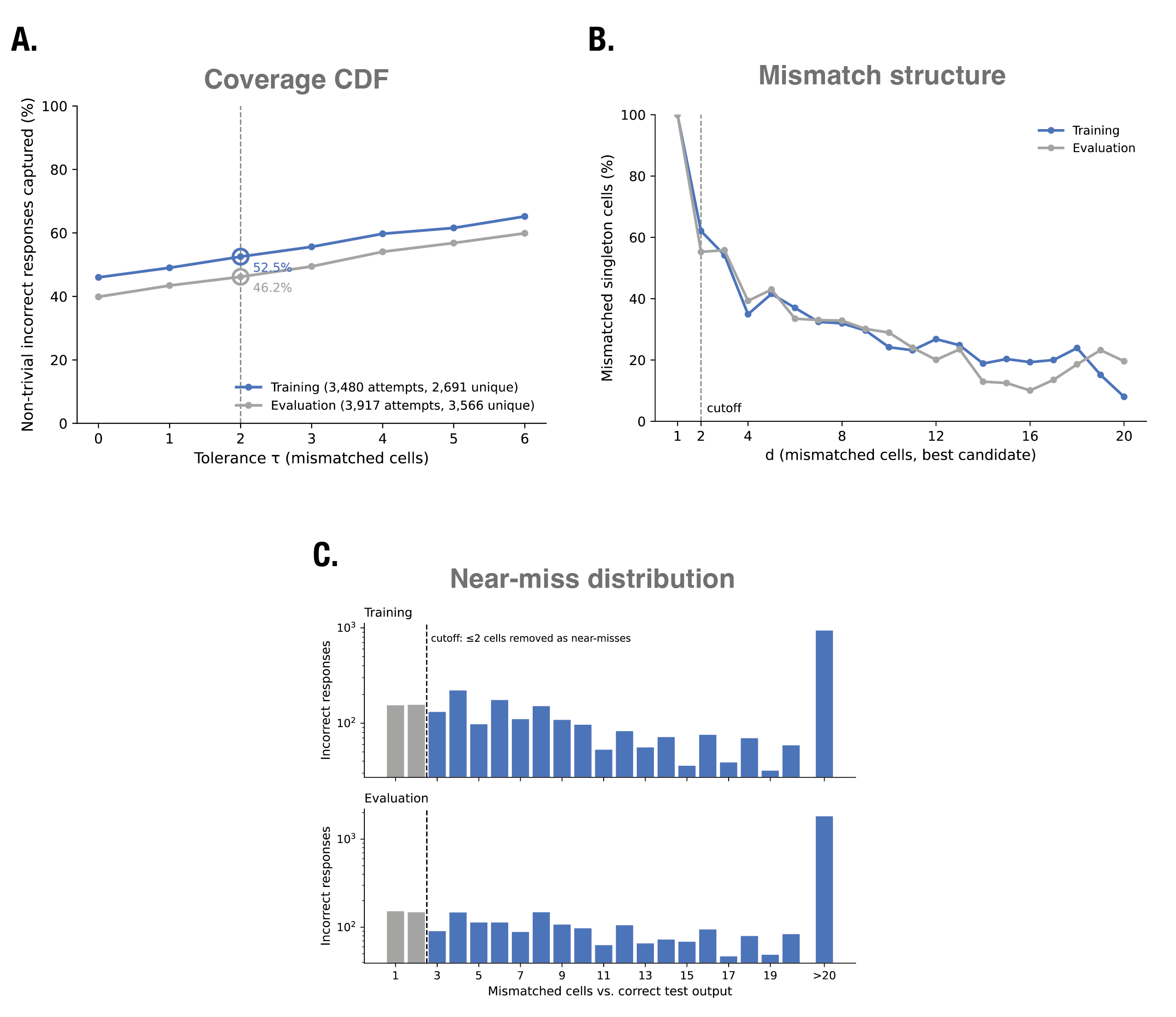}
    \caption{\textbf{The two-cell cutoff separates procedural noise from rule-level divergence.} (\textbf{A}) Coverage of non-trivial incorrect responses as a function of the tolerance $\tau$, attempt-weighted, shown separately for the \texttt{training} (blue) and \texttt{evaluation} (grey) splits. Circles mark the value $\tau = 2$ used throughout ($52.5\%$ and $46.2\%$ respectively). (\textbf{B}) Share of mismatched cells that are scattered singletons---no mismatched cell among their four neighbors---as a function of $d$, the mismatch count of the best-scoring candidate program. The steep fall immediately past the cutoff shows that small residuals are isolated misclick-like cells, whereas higher values of $d$ indicate structured regions/objects. (\textbf{C}) Distribution of cell-level distances between incorrect responses and the correct test output, per split. Grey bars ($\leq 2$ mismatched cells) are the near-misses removed by the filter.}\label{fig:sup_coverage}
\end{figure}

  Supplementary Figure~\ref{fig:sup_coverage} quantifies the same point over the full dataset. Panel A gives the coverage of incorrect attempts as a function of the tolerance $\tau$: coverage is the empirical CDF of $d$, the minimum
  number of mismatched cells between any of the fifteen sampled programs for an error and the target response, restricted to size-matched outputs. At $\tau = 2$, coverage is $52.5\%$ on the \verb|training| split and $46.2\%$ on the \verb|evaluation| split. The curve keeps climbing beyond the cutoff, so the threshold is conservative. Panel B shows that the share of mismatched cells that are scattered singletons (no mismatched neighbor) falls from $\sim\!60\%$ at $d = 2$ to $\sim\!35\%$ by $d = 4$ and to $\sim\!20\%$ beyond $d = 10$. Residuals of three or more cells are therefore likely to represent objects or other structures, but more analyses are needed to confirm this. In any case, accepting them would include programs implementing a different transformation, as opposed to a slightly imprecise one. Panel C shows the same cutoff from the filtering side: the distribution of cell-level distances between incorrect responses and the correct test output, with the removed near-miss data ($\leq 2$ cells) shaded.

  Supplementary Table~\ref{tab:sup_filtering} reports what the trivial-error filter removes.

\clearpage
\enlargethispage{35mm}
\clearpage

\clearpage
\enlargethispage{35mm}
\clearpage

  \input{tables/sup_filtering.tex}

  \FloatBarrier

  \section{Modeling coverage and program yield by split}\label{sup:coverage}

  Supplementary Table~\ref{tab:sup_coverage} gives the per-split decomposition of the coverage results reported pooled in \S\ref{sec:results-coverage}. \verb|training| reference programs are
  hand-written whereas \verb|evaluation| reference programs are
  model-synthesized (\S\ref{sec:synthetic-solvers}). Edits
  were sampled with \verb|Gemini 3 Pro| on \verb|training| and
  \verb|Gemini 3.5 Flash| on \verb|evaluation|.

  \input{tables/sup_coverage.tex}

  \FloatBarrier

  \section{Rule types and order effects by split}\label{sup:ruletypes}

  Supplementary Table~\ref{tab:sup_ruletypes} reports the per-split values behind
  the pooled rule-type and order-effect estimates of
  \S\ref{sec:results-partial}. Estimates are attempt-weighted population
  marginals from the hierarchical categorical model
  (\S\ref{sec:rule-type-model}). The prevalence of programs consistent with at least one training example is close across splits ($23.8\%$ versus $21.3\%$), whereas the structured share of partial matches is substantially higher on \verb|training|
  ($62.5\%$ versus $48.5\%$), with the largest proportion coming from the first-$k$ tier.

  \input{tables/sup_ruletypes.tex}

  \FloatBarrier

  \section{Program length, reference complexity and error categories}\label{sup:length}

\begin{figure}[h]
    \centering
    \includegraphics[width=0.95\textwidth]{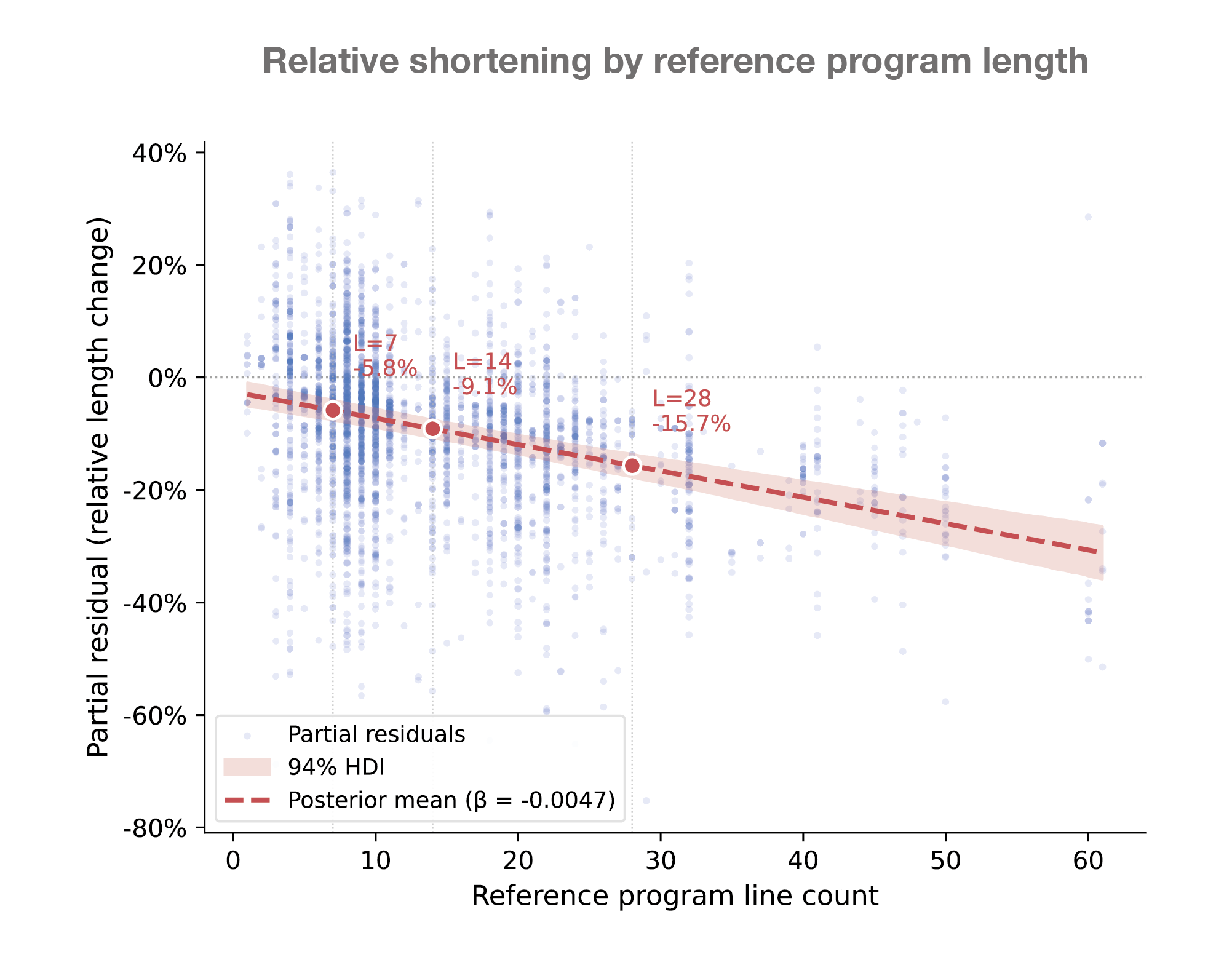}
    \caption{\textbf{Relative length change as a function of reference complexity and error category.} Partial residuals of relative length change against reference-program line count, with the posterior mean slope (dashed) and its 94\% HDI (band). Annotations mark the predicted shortening at $L = 7$, $14$ and $28$ lines.}\label{fig:sup_length}
  \end{figure}

  Supplementary Figure~\ref{fig:sup_length} and Supplementary
  Table~\ref{tab:sup_rlc_category} report the full output of the program length model of \S\ref{sec:error-cat-diff-model}, which regresses relative length change on reference-program line count and task success rate with a soft error-category term.

  Beyond the growing relative shortening predicted as a function of target rule complexity (\S\ref{sec:results-simplicity}), our results suggest substantial variation in relative shortening across error categories (Supplementary Table~\ref{tab:sup_rlc_category}). Around a grand-mean shortening of $9.2\%$ (94\% HDI $[7.6, 10.8]$), the seven cognitive categories span roughly $34$ percentage points. At one end, partial-program errors produce the greatest simplification with programs $32.3\%$ shorter than the target rule (94\% HDI $[29.1, 35.6]$; $\gamma = -0.231$, 94\% HDI $[-0.264, -0.198]$). Conversely, boundary precision errors yield no reliable simplification: although they deviate credibly from the grand mean ($\gamma = +0.111$, 94\% HDI $[0.065, 0.159]$), their predicted length is $2.0\%$ longer than the reference program, with an interval that includes zero (94\% HDI $[-3.0, 7.3]$).

  Overall, we find that our analysis recovers an interpretable hierarchy between the categories. This result suggests that when people make simple localized errors such as boundary precision or operator confusion, the rules they infer, as modeled by our programs, approximate much of the structure of the target rule. Conversely, when people only partially execute the rule or select the wrong transformation, their approximations are coarser, which manifests as substantially shorter programs.

  \input{tables/sup_rlc_category.tex}

  \FloatBarrier

\section{Machine errors}\label{sup:llm_alltasks}

\begin{figure}[p]
    \centering
    \includegraphics[width=\textwidth,keepaspectratio]{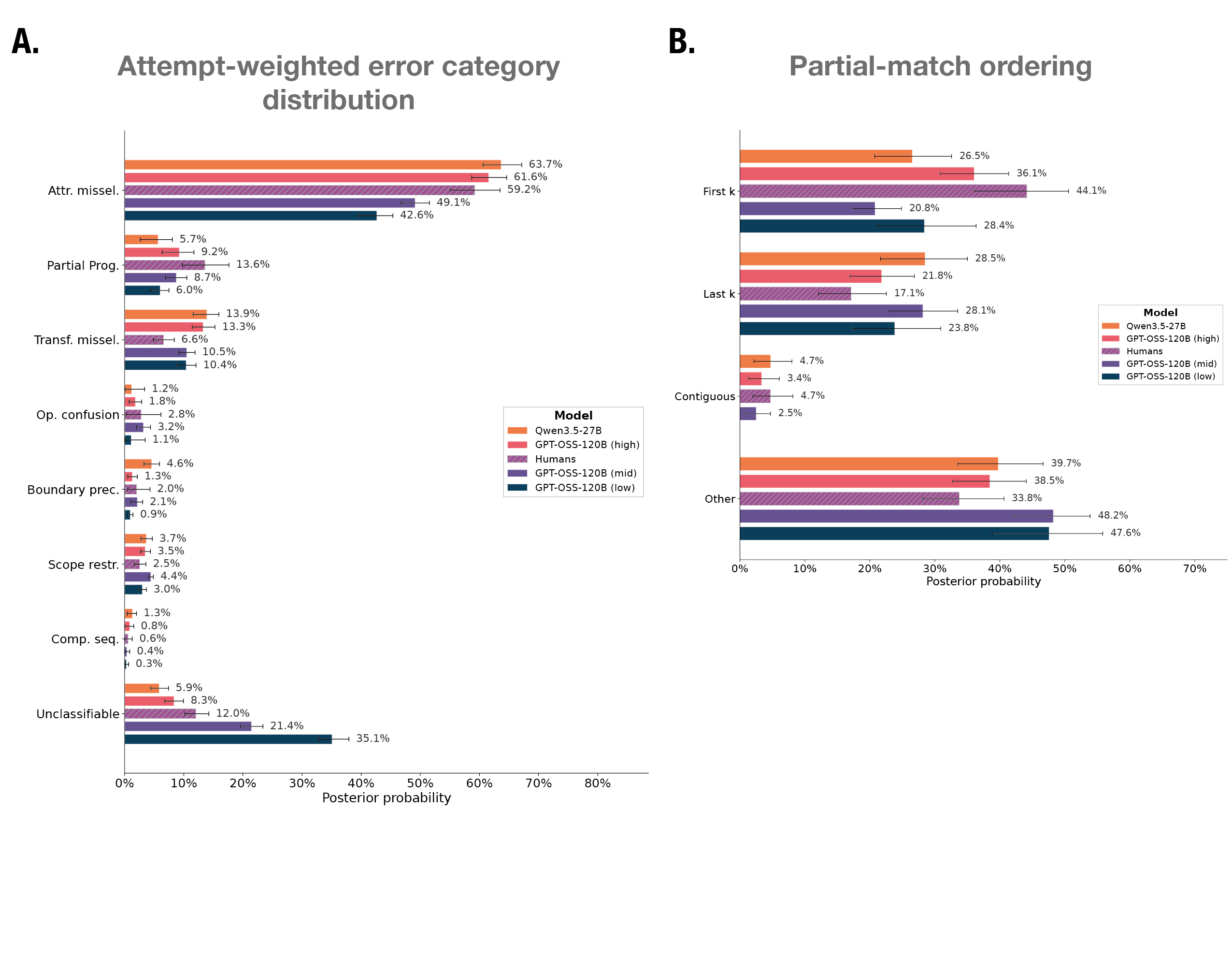}
    \caption{\textbf{Language models display human-like error category profiles, but order effects are weaker and less consistent than in humans.} All panels report values corresponding to the subset of 56 tasks where evaluated LRMs and people both generated at least one incorrect response. Human responses are restricted to first-attempt comparisons. (\textbf{A}) Prevalence of each error category among modeled incorrect machine errors. As with human results, an LLM classifier assigned every program one of seven cognitive categories or \textit{unclassifiable}, conditioned on the reference program, the task grids, the synthesized program and its execution trace (Table~\ref{tab:error-taxonomy} for definitions). Bars give the posterior median of the attempt-weighted prevalence derived from a hierarchical Bayesian categorical model, and error bars the 94\% HDI. The categories are mutually exclusive. (\textbf{B}) Conditional on an attempt being explained by a partial-match program, we show the positions of the training examples it reproduces: the first-k or last-k example(s) shown, a contiguous block of at least two examples that does not include the first or last examples, or any other configuration. In both plots, bars give the posterior median attempt-weighted prevalence and error bars the 94\% HDI.}\label{fig:sup_llm_matched}
\end{figure}

Supplementary Table~\ref{tab:sup_llm_filtering} reports what the trivial-error filter removes from each model configuration's incorrect responses before modeling, as the counterpart to Supplementary Table~\ref{tab:sup_filtering} for human responses. Supplementary Tables~\ref{tab:sup_llm_yield}--\ref{tab:sup_llm_categories} then report the extended LRM evaluation and associated error analyses on the full set of errors for comparison with the restricted-subset results reported in the main text. Figure~\ref{fig:sup_llm_matched} provides additional visualization of error analyses on the subset of tasks where both models and people produced modeled incorrect responses. Supplementary Table~\ref{tab:sup_llm_source_prediction} reports within-task and pooled ROC-AUC for the human-versus-LRM source classifiers on this same subset and for each classifier and Supplementary Table~\ref{tab:sup_llm_source_prediction_by_config} reports classifiers trained on humans versus each LRM configuration separately (\S\ref{sec:methods-source-prediction}).

    \input{tables/sup_llm_filtering.tex}
    \input{tables/sup_llm_yield.tex}
    \input{tables/sup_llm_ruletypes.tex}
    \input{tables/sup_llm_categories.tex}
    \input{tables/sup_llm_source_prediction.tex}
    \input{tables/sup_llm_source_prediction_by_config.tex}

\clearpage
\enlargethispage{35mm}
\clearpage

  \FloatBarrier

  \section{Prompts}\label{sup:prompts}

\begin{figure}[p]
    \centering
    \includegraphics[width=\textwidth,keepaspectratio]{\detokenize{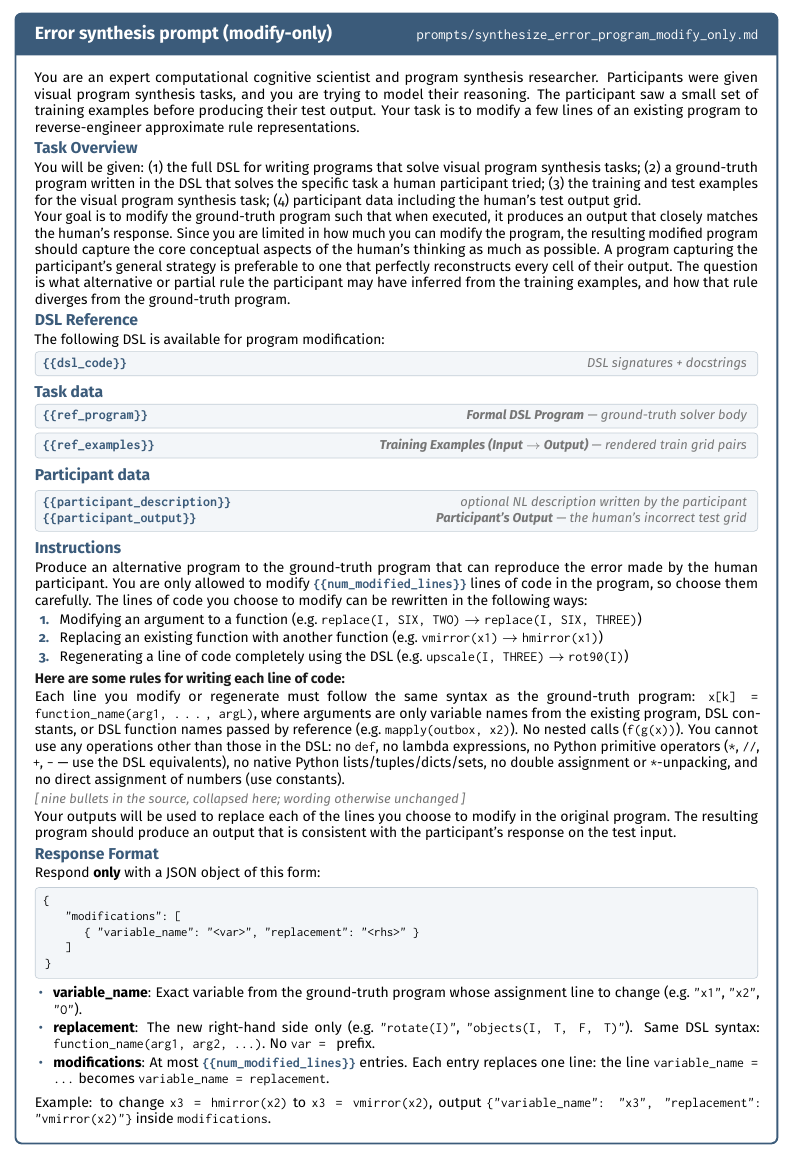}}
    \caption{\textbf{Error-program synthesis prompt (\S\ref{sec:methods-pipeline}).}
      The model is given the full DSL specification, the task grids, the reference
      program implementing the target rule, and the respondent's incorrect output
      grid, and returns a structured set of line edits under a fixed JSON
      schema.}\label{fig:sup_prompt_synthesis}
  \end{figure}

\begin{figure}[p]
    \centering
    \includegraphics[width=\textwidth,keepaspectratio]{\detokenize{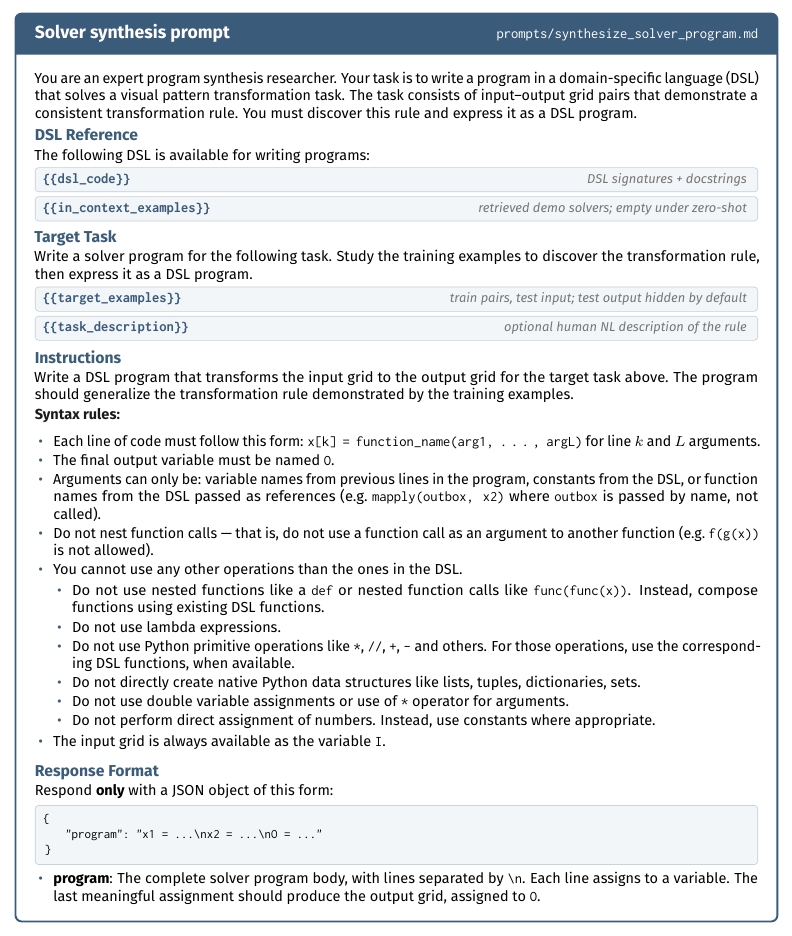}}
    \caption{\textbf{Reference-program (solver) synthesis prompt
        (\S\ref{sec:synthetic-solvers}).} Used to generate reference programs for
      \texttt{evaluation}-split tasks, which have no hand-written solver. The
      prompt is retrieval-augmented with five solved tasks and their human-written
      solvers from the \texttt{training} split.}\label{fig:sup_prompt_solver}
  \end{figure}

\begin{figure}[p]
    \centering
    \includegraphics[width=\textwidth,keepaspectratio]{\detokenize{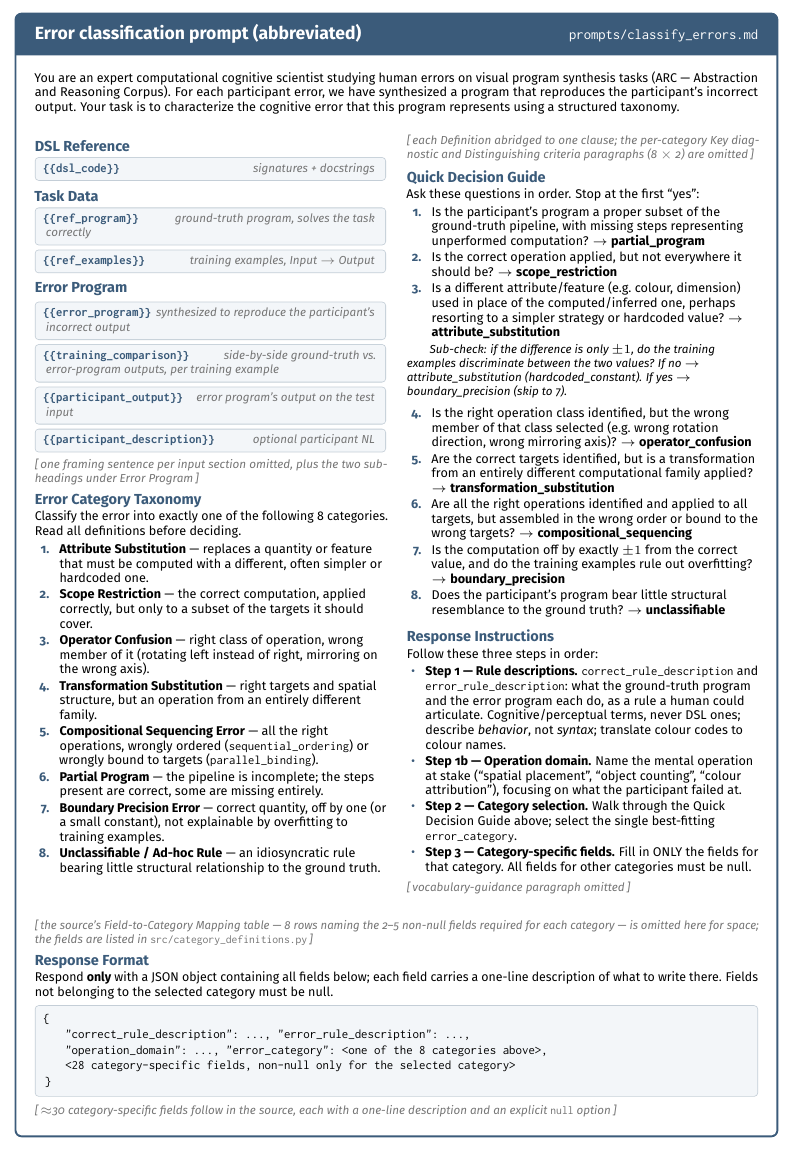}}
    \caption{\textbf{Error-classification prompt (\S\ref{sec:methods-classification}).}
      The classifier sees the reference program, the task grids, the synthesized
      error program, and a side-by-side comparison of their behavior on the
      training examples, and returns one of the seven cognitive categories or the
      unclassifiable catch-all.}\label{fig:sup_prompt_classification}
  \end{figure}

\begin{figure}[p]
    \centering
    \includegraphics[width=\textwidth,keepaspectratio]{\detokenize{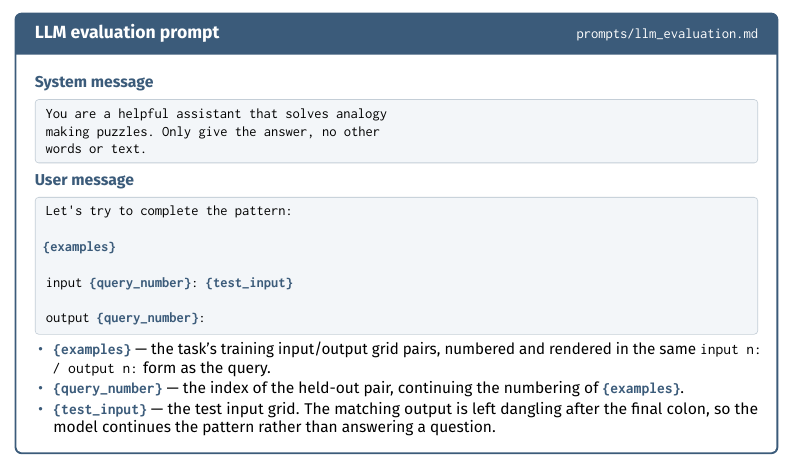}}
    \caption{\textbf{Evaluation prompt (\S\ref{sec:methods-llms}).} Used to collect machine responses on ARC tasks for comparison with the human data. The model is shown the training pairs and asked to complete the held-out pair.
    }\label{fig:sup_prompt_llm_eval}
  \end{figure}

  Supplementary Figures~\ref{fig:sup_prompt_synthesis}--\ref{fig:sup_prompt_llm_eval}
  reproduce the three prompt templates used by the pipeline. Placeholders in
  double braces are substituted at run time with the corresponding task or
  program text.

  \FloatBarrier

\bibliography{main-bibliography}%

\end{document}

%% file: tables/error_taxonomy.tex
\begin{table}[t]
  \centering
  \caption{\textbf{Taxonomy of error categories used to classify synthesized programs.} Each category identifies a way in which programs that reproduce incorrect responses deviate from the reference program implementing the target rule for the same task. Categories are defined over structural differences between the two programs, but the classifier is also conditioned on the response grids. Categories are defined as mutually exclusive and the ``unclassifiable'' catch-all makes the set exhaustive. The examples in the third column are schematic illustrations of the contrast between reference and response programs that defines each category. Every synthesized program was classified by an LLM given a reference program, task grids, the synthesized program, and its execution trace on the training and test inputs, with five independent classifications sampled per program (\S\ref{sec:methods-classification}).}

  \label{tab:error-taxonomy}
  \footnotesize
  \renewcommand{\arraystretch}{1.15}
  \begin{tabularx}{\linewidth}{@{}l X l@{}}
    \toprule
    \textbf{Category}           & \textbf{Definition}                                                                                                   & \textbf{Example}                       \\
    \midrule
    Attribute misselection      & The correct operation is applied, but with a misselected attribute, parameter, or input value.                        & inferred color $\to$ hardcoded color \\
    Scope restriction           & The correct computation is applied, but only to a subset of the entities or regions where it should apply. & all objects $\to$ subset               \\
    Operator confusion          & The correct class of operation is identified, but the wrong member within that class is selected.                     & rotate CW $\to$ rotate CCW             \\
    Transformation misselection & The correct targets are identified, but a transformation from an entirely different computational family is applied.  & mirror $\to$ recolor                  \\
    Sequencing                  & All required operations are identified, but they are composed in the wrong order or bound to the wrong targets.       & ($A$ then $B$) $\to$ ($B$ then $A$)    \\
    Partial program             & A subset of the required steps are executed correctly; one or more computational blocks are entirely missing.         & extract $+$ tile $\to$ extract only    \\
    Boundary precision          & The correct quantity is computed from the correct source, but with a small additive offset (off-by-one).              & $h \to h-1$                            \\
    Unclassifiable              & The response reflects an idiosyncratic rule with no recognizable structural relationship to the reference program.    & ad-hoc rule                            \\
    \bottomrule
  \end{tabularx}
\end{table}

%% file: tables/sup_filtering.tex
\begin{table}[!ht]
  \centering
  \caption{\textbf{Trivial-error filtering by split (\S\ref{sec:methods-data}).} Counts and percentages of incorrect responses removed before modeling, under the noise tolerance criterion: i) shapes must match, ii) grids of ten or more cells tolerate up to two mismatched cells, iii) smaller grids must match exactly and iv) copies of a training example are removed only on an exact match at any size. Percentages are over the full pre-filter pool of responses of each split.}\label{tab:sup_filtering}
  \footnotesize
  \begin{tabular}{@{}lrrr@{}}
    \toprule
     & \textbf{Training} & \textbf{Evaluation} & \textbf{Combined} \\
    \midrule
    \multicolumn{4}{@{}l}{\emph{By incorrect responses (attempts)}}\\
    Incorrect responses & 4{,}293 & 5{,}156 & 9{,}449 \\
    Filtered (all trivial) & 808 (18.8\%) & 891 (17.3\%) & 1{,}699 (18.0\%) \\
    \quad Near-miss ($\approx$ correct) & 189 (4.4\%) & 251 (4.9\%) & 440 (4.7\%) \\
    \quad Test-input copy (identity) & 368 (8.6\%) & 377 (7.3\%) & 745 (7.9\%) \\
    \quad Training-example copy & 130 (3.0\%) & 197 (3.8\%) & 327 (3.5\%) \\
    \quad Empty $3\times3$ grid & 121 (2.8\%) & 66 (1.3\%) & 187 (2.0\%) \\
    \midrule
    \multicolumn{4}{@{}l}{\emph{By unique incorrect grids}}\\
    Unique incorrect grids & 3{,}220 & 4{,}589 & 7{,}809 \\
    Filtered (all trivial) & 525 (16.3\%) & 697 (15.2\%) & 1{,}222 (15.6\%) \\
    \quad Near-miss ($\approx$ correct) & 140 (4.3\%) & 213 (4.6\%) & 353 (4.5\%) \\
    \quad Test-input copy (identity) & 209 (6.5\%) & 282 (6.1\%) & 491 (6.3\%) \\
    \quad Training-example copy & 88 (2.7\%) & 142 (3.1\%) & 230 (2.9\%) \\
    \quad Empty $3\times3$ grid & 88 (2.7\%) & 60 (1.3\%) & 148 (1.9\%) \\
    \bottomrule
  \end{tabular}
\end{table}

%% file: tables/sup_coverage.tex
\begin{table}[!ht]
  \centering
  \caption{\textbf{Modeling coverage and program yield by split.} Unless specified, all reported numbers refer to errors modeled under the two-cell tolerance criterion. Attempt-weighted quantities weight each unique error by the number of participants who produced it. Exact-match reproduced is the percentage of errors with at least one program reproducing the grid with zero mismatched cells. Effective edits are counted after the parser removes causally irrelevant lines. Four training-split errors were missing from the synthesis run, so program samples cover 2,691 of the 2,695 training errors.}\label{tab:sup_coverage}
  \footnotesize
  \begin{tabular}{@{}lrrr@{}}
    \toprule
     & \textbf{Training} & \textbf{Evaluation} & \textbf{Combined} \\
    \midrule
    Reference program & hand-written & synthesized & --- \\
    Proposal model & Gemini 3 Pro & Gemini 3.5 Flash & --- \\
    Tasks modeled & 362 & 361 & 723 \\
    Unique non-trivial errors & 2{,}695 & 3{,}566 & 6{,}261 \\
    Incorrect attempts & 3{,}485 & 3{,}917 & 7{,}402 \\
    \midrule
    Program samples collected & 40{,}365 & 53{,}490 & 93{,}855 \\
    Tasks with $\geq 1$ error modeled & 304 & 304 & 608 \\
    Errors reproduced & 1{,}274 (47.3\%) & 1{,}507 (42.3\%) & 2{,}781 (44.4\%) \\
    Attempts reproduced & 1{,}827 (52.4\%) & 1{,}809 (46.2\%) & 3{,}636 (49.1\%) \\
    Participants modeled & 548 & 611 & 1{,}159 \\
    Exact-match coverage (errors) & 39.9\% & 35.5\% & 37.4\% \\
    Exact-match coverage (attempts) & 45.9\% & 39.9\% & 42.7\% \\
    \midrule
    Total filtered programs & 3{,}690 & 4{,}556 & 8{,}246 \\
    Programs per error (M $\pm$ SD) & 2.90 $\pm$ 2.04 & 3.02 $\pm$ 2.01 & 2.97 $\pm$ 2.02 \\
    Programs per error (median [IQR]) & 2 [1, 4] & 3 [1, 4] & 2 [1, 4] \\
    Programs per error (max) & 10 & 12 & 12 \\
    Effective edits / program (M, SD) & 3.30, 2.00 & 3.63, 2.16 & 3.49, 2.10 \\
    Effective edits, per-error min. (M $\pm$ SD) & 2.69 $\pm$ 1.97 & 2.96 $\pm$ 2.09 & 2.84 $\pm$ 2.04 \\
    \bottomrule
  \end{tabular}
\end{table}

%% file: tables/sup_ruletypes.tex
\begin{table}[!ht]
  \centering
  \caption{\textbf{Rule types and order effects by split (attempt-weighted).} Posterior median and 94\% HDI. A program is a \emph{partial match} if it reproduces at least one but not all training outputs, an \emph{alternative rule} if it reproduces all of them yet still fails the test example, and \emph{neither} otherwise. Order effects are conditional on a partial match and describe which of the training examples the program reproduces, ordered as models or participants saw them.}\label{tab:sup_ruletypes}
  \footnotesize
  \begin{tabular}{@{}lccc@{}}
    \toprule
     & \textbf{Training} & \textbf{Evaluation} & \textbf{Combined} \\
    \midrule
    \emph{n} (attempts / errors) & 1{,}827 / 1{,}274 & 1{,}809 / 1{,}507 & 3{,}636 / 2{,}781 \\
    \midrule
    \multicolumn{4}{@{}l}{\emph{Rule type (all modeled attempts)}}\\
    Neither & 76.2\% [74.9, 77.4] & 78.7\% [77.8, 79.6] & 77.4\% [76.6, 78.1] \\
    Partial match & 22.1\% [20.8, 23.3] & 19.7\% [18.8, 20.6] & 20.9\% [20.2, 21.7] \\
    Alternative rule & 1.8\% [1.3, 2.3] & 1.6\% [1.2, 2.1] & 1.7\% [1.4, 2.0] \\
    \textbf{Consistent with $\geq 1$ example} & \textbf{23.8\% [22.6, 25.1]} & \textbf{21.3\% [20.4, 22.2]} & \textbf{22.6\% [21.9, 23.4]} \\
    \midrule
    \multicolumn{4}{@{}l}{\emph{Order effects (conditional on a partial match)}}\\
    \emph{n} (attempts / errors) & 507 / 302 & 472 / 311 & 979 / 613 \\
    First-$k$ (prefix run) & 41.0\% [38.2, 44.0] & 30.1\% [27.6, 32.6] & 35.8\% [34.0, 37.7] \\
    Last-$k$ (suffix run) & 17.0\% [14.8, 19.4] & 15.7\% [13.8, 17.8] & 16.4\% [14.9, 17.8] \\
    Contiguous (interior run $> 1$) & 4.4\% [2.9, 5.9] & 2.6\% [1.6, 3.7] & 3.5\% [2.6, 4.5] \\
    Other (single interior or scattered) & 37.5\% [34.4, 40.6] & 51.5\% [48.7, 54.4] & 44.3\% [42.2, 46.3] \\
    \textbf{Structured subset} & \textbf{62.5\% [59.4, 65.6]} & \textbf{48.5\% [45.6, 51.3]} & \textbf{55.7\% [53.7, 57.8]} \\
    \bottomrule
  \end{tabular}
\end{table}

%% file: tables/sup_rlc_category.tex
\begin{table}[!ht]
  \centering
  \caption{\textbf{Error-category effects on relative program length (\texttt{training} split).} $\gamma_m$ is the deviation from the grand mean of the seven cognitive categories, which share a sum-to-zero prior; negative $\gamma$ means more relative shortening than the average error type. The \emph{unclassifiable} catch-all ($\dagger$) is held outside that constraint. ``\% shorter'' is the absolute per-category prediction $\beta_0 + \gamma_m$ at mean success rate and mean reference length, summed per posterior draw (grand-mean shortening $\beta_0 = 9.2\%$); a negative value means the error programs are \emph{longer} than the reference. $n = 3{,}690$ programs over 1{,}274 errors.}\label{tab:sup_rlc_category}
  \footnotesize
  \begin{tabular}{@{}lccc@{}}
    \toprule
    \textbf{Category} & \textbf{$\gamma$ [94\% HDI]} & \textbf{$P(>0)$} & \textbf{\% shorter [94\% HDI]} \\
    \midrule
    Unclassifiable $\dagger$ & $-0.369$ [$-0.395$, $-0.344$] & 0.000 & $+46.1$ [$+44.2$, $+48.0$] \\
    Partial program & $-0.231$ [$-0.264$, $-0.198$] & 0.000 & $+32.3$ [$+29.1$, $+35.6$] \\
    Transformation misselection & $-0.121$ [$-0.150$, $-0.091$] & 0.000 & $+21.3$ [$+18.4$, $+24.0$] \\
    Attribute misselection & $+0.014$ [$-0.007$, $+0.034$] & 0.893 & $+7.8$ [$+6.2$, $+9.3$] \\
    Scope restriction & $+0.045$ [$-0.006$, $+0.098$] & 0.947 & $+4.6$ [$-1.1$, $+10.4$] \\
    Operator confusion & $+0.080$ [$+0.041$, $+0.118$] & 1.000 & $+1.2$ [$-2.8$, $+5.1$] \\
    Sequencing & $+0.102$ [$+0.039$, $+0.164$] & 0.999 & $-1.1$ [$-7.9$, $+6.0$] \\
    Boundary precision & $+0.111$ [$+0.065$, $+0.159$] & 1.000 & $-2.0$ [$-7.3$, $+3.0$] \\
    \bottomrule
  \end{tabular}
\end{table}

%% file: tables/sup_llm_filtering.tex
\begin{table}[!ht]
  \centering
  \caption{\textbf{Trivial-error filtering by model configuration (\S\ref{sec:methods-data}).} Incorrect model responses with a valid grid removed before modeling, under the same noise tolerance criterion applied to human responses (Supplementary Table~\ref{tab:sup_filtering}), counted by response (top) and by unique incorrect grid (bottom). Percentages are over each configuration's pre-filter total in the same block.}\label{tab:sup_llm_filtering}
  \scriptsize
  \setlength{\tabcolsep}{2pt}
  \begin{tabular}{@{}lrrrr@{}}
    \toprule
     & \textbf{Qwen3.5-27B} & \multicolumn{3}{c}{\textbf{GPT-OSS-120B}} \\
    \cmidrule(lr){3-5}
     &  & \textbf{high} & \textbf{mid} & \textbf{low} \\
    \midrule
    \multicolumn{5}{@{}l}{\emph{By incorrect responses (attempts)}}\\
    Responses with a valid grid & 689 & 742 & 2{,}002 & 3{,}059 \\
    Filtered (all trivial) & 114 (16.5\%) & 90 (12.1\%) & 264 (13.2\%) & 568 (18.6\%) \\
    \quad Near-miss ($\approx$ correct) & 73 (10.6\%) & 54 (7.3\%) & 84 (4.2\%) & 61 (2.0\%) \\
    \quad Test-input copy (identity) & 22 (3.2\%) & 12 (1.6\%) & 110 (5.5\%) & 354 (11.6\%) \\
    \quad Training-example copy & 19 (2.8\%) & 24 (3.2\%) & 70 (3.5\%) & 153 (5.0\%) \\
    \midrule
    \multicolumn{5}{@{}l}{\emph{By unique incorrect grids}}\\
    Unique incorrect grids & 536 & 562 & 1{,}522 & 2{,}059 \\
    Filtered (all trivial) & 79 (14.7\%) & 57 (10.1\%) & 148 (9.7\%) & 229 (11.1\%) \\
    \quad Near-miss ($\approx$ correct) & 57 (10.6\%) & 42 (7.5\%) & 72 (4.7\%) & 47 (2.3\%) \\
    \quad Test-input copy (identity) & 13 (2.4\%) & 7 (1.2\%) & 48 (3.2\%) & 110 (5.3\%) \\
    \quad Training-example copy & 9 (1.7\%) & 8 (1.4\%) & 28 (1.8\%) & 72 (3.5\%) \\
    \bottomrule
  \end{tabular}
\end{table}

%% file: tables/sup_llm_yield.tex
\begin{table}[!ht]
  \centering
  \caption{\textbf{Evaluation and modeling yield by model configuration (across all tasks with errors).} Accuracy is the mean over ten seeds on the 400 \texttt{training} tasks of ARC-AGI-1. Attempts correspond to every incorrect response that survived filtering; unique errors deduplicate identical grid outputs. Attempts modeled and errors modeled are the counts with at least one verified program reproducing the error within the two-cell noise tolerance criterion---with coverage as a percentage of attempts and of unique errors, respectively.}\label{tab:sup_llm_yield}
  \scriptsize
  \setlength{\tabcolsep}{2pt}
  \begin{tabular}{@{}lrrrr@{}}
    \toprule
     & \textbf{Qwen3.5-27B} & \multicolumn{3}{c}{\textbf{GPT-OSS-120B}} \\
    \cmidrule(lr){3-5}
     &  & \textbf{high} & \textbf{mid} & \textbf{low} \\
    \midrule
    Accuracy (\%) & 80.63 & 72.28 & 50.96 & 25.02 \\
    Tasks with $\geq 1$ error & 145 & 173 & 284 & 344 \\
    Incorrect responses & 806 & 1{,}153 & 2{,}040 & 3{,}119 \\
    \quad with a valid grid & 689 & 742 & 2{,}002 & 3{,}059 \\
    Attempts & 575 & 652 & 1{,}738 & 2{,}491 \\
    Unique errors & 457 & 505 & 1{,}374 & 1{,}830 \\
    \midrule
    Program samples & 8{,}625 & 9{,}780 & 26{,}070 & 37{,}365 \\
    Attempts modelled & 296 (51.5\%) & 355 (54.4\%) & 1{,}031 (59.3\%) & 1{,}565 (62.8\%) \\
    Errors modelled & 197 (43.1\%) & 236 (46.7\%) & 714 (52.0\%) & 991 (54.2\%) \\
    Filtered programs & 823 & 944 & 2{,}760 & 3{,}804 \\
    \bottomrule
  \end{tabular}
\end{table}

%% file: tables/sup_llm_ruletypes.tex
\begin{table}[!ht]
  \centering
  \caption{\textbf{Simplicity bias, rule types and order effects by model configuration (across all tasks with errors).} $\beta$ is the posterior mean relative shortening of error programs against correct-output controls, where negative values indicate shorter programs. Rule types and order effects are attempt-weighted population marginals from the hierarchical categorical model (\S\ref{sec:rule-type-model}); posterior median and 94\% HDI, in percent. A program is a \emph{partial match} if it reproduces at least one but not all training outputs, an \emph{alternative rule} if it reproduces all of them yet still fails the test example, and \emph{neither} otherwise. Order effects are conditional on a partial match and describe which training examples the program reproduces, ordered as models or participants saw them.}\label{tab:sup_llm_ruletypes}
  \scriptsize
  \setlength{\tabcolsep}{2pt}
  \begin{tabular}{@{}lcccc@{}}
    \toprule
     & \textbf{Qwen3.5-27B} & \multicolumn{3}{c}{\textbf{GPT-OSS-120B}} \\
    \cmidrule(lr){3-5}
     &  & \textbf{high} & \textbf{mid} & \textbf{low} \\
    \midrule
    \multicolumn{5}{@{}l}{\emph{Simplicity bias (relative length change vs.\ correct-output control)}}\\
    $\beta$ & $-0.147$ & $-0.153$ & $-0.211$ & $-0.245$ \\
    \quad 94\% HDI & [$-0.185$,$-0.112$] & [$-0.187$,$-0.119$] & [$-0.236$,$-0.185$] & [$-0.268$,$-0.221$] \\
    $P(\beta < 0)$ & 1.00 & 1.00 & 1.00 & 1.00 \\
    \midrule
    \multicolumn{5}{@{}l}{\emph{Rule type (all modeled attempts)}}\\
    Neither & 59.1 [56.5,61.8] & 60.2 [57.8,62.8] & 78.6 [77.2,79.8] & 84.2 [83.2,85.4] \\
    Partial match & 33.6 [31.0,36.2] & 33.0 [30.5,35.4] & 19.6 [18.3,20.8] & 14.7 [13.6,15.7] \\
    Alternative rule & 7.3 [5.6,8.9] & 6.7 [5.3,8.2] & 1.8 [1.3,2.3] & 1.1 [0.7,1.5] \\
    \midrule
    \multicolumn{5}{@{}l}{\emph{Order effects (conditional on a partial match)}}\\
    \emph{n} (errors) & 146 & 160 & 291 & 388 \\
    First-$k$ (prefix run) & 28.7 [24.0,33.7] & 41.3 [37.2,45.6] & 40.1 [36.8,43.2] & 37.6 [34.1,41.2] \\
    Last-$k$ (suffix run) & 27.9 [22.4,33.0] & 19.7 [15.3,23.6] & 25.4 [22.2,28.6] & 26.2 [22.9,29.6] \\
    Contiguous (interior) & 3.8 [1.9,6.4] & 2.3 [0.9,4.1] & 1.3 [0.4,2.3] & 1.7 [0.7,3.0] \\
    Other & 39.3 [34.0,45.0] & 36.5 [32.0,41.1] & 33.2 [29.6,36.7] & 34.4 [30.4,38.0] \\
    \bottomrule
  \end{tabular}
\end{table}

%% file: tables/sup_llm_categories.tex
\begin{table}[!ht]
  \centering
  \caption{\textbf{Error categories by model configuration (across all tasks with errors).} Attempt-weighted population marginals from the hierarchical categorical model (\S\ref{sec:error-class-model}); posterior median and 94\% HDI, in percent. Error categories are the seven cognitive types of \S\ref{sec:results-taxonomy} and \emph{unclassifiable} ($\dagger$) is the catch-all category. \emph{n} is the number of errors with at least one validly classified program; one medium-effort and two low-effort errors (7 and 3 programs) had no valid classification sample and are excluded, hence 713/989 rather than 714/991 in Supplementary Table~\ref{tab:sup_llm_yield}.}\label{tab:sup_llm_categories}
  \scriptsize
  \setlength{\tabcolsep}{2pt}
  \begin{tabular}{@{}lcccc@{}}
    \toprule
     & \textbf{Qwen3.5-27B} & \multicolumn{3}{c}{\textbf{GPT-OSS-120B}} \\
    \cmidrule(lr){3-5}
     &  & \textbf{high} & \textbf{mid} & \textbf{low} \\
    \midrule
    \emph{n} (errors) & 197 & 236 & 713 & 989 \\
    \midrule
    Attribute misselection & 61.0 [58.7,63.5] & 58.4 [55.9,60.9] & 45.7 [44.3,47.3] & 38.0 [36.6,39.4] \\
    Transformation misselection & 13.0 [11.4,14.5] & 12.6 [11.1,14.3] & 11.0 [10.0,12.1] & 10.1 [9.3,10.9] \\
    Partial program & 7.2 [5.4,8.8] & 10.2 [8.2,12.2] & 10.8 [9.6,12.0] & 9.5 [8.4,10.6] \\
    Scope restriction & 3.2 [2.3,4.3] & 3.9 [3.0,4.7] & 2.7 [2.4,3.1] & 3.0 [2.5,3.6] \\
    Operator confusion & 0.9 [0.1,2.3] & 3.2 [2.0,4.7] & 2.7 [1.9,3.5] & 3.6 [2.7,4.5] \\
    Boundary precision & 3.4 [2.4,4.4] & 1.3 [0.6,2.0] & 1.8 [1.3,2.3] & 1.5 [1.1,1.8] \\
    Sequencing & 1.6 [0.8,2.3] & 0.5 [0.0,1.1] & 2.3 [1.8,2.9] & 0.9 [0.5,1.3] \\
    Unclassifiable $\dagger$ & 9.6 [8.2,10.9] & 9.7 [8.3,11.2] & 22.8 [21.8,24.1] & 33.4 [32.3,34.5] \\
    \bottomrule
  \end{tabular}
\end{table}

%% file: tables/sup_llm_source_prediction.tex
\begin{table}[!ht]
  \centering
  \caption{\textbf{Human-versus-LLM source prediction (\S\ref{sec:methods-source-prediction}).} Out-of-fold ROC-AUC on the 811 errors (144 human, 667 LLM) from the 56 matched tasks, for classifiers trained on primitive delta vectors from the errors on all tasks (651 human, 2{,}138 LLM). LLM errors are the positive class. Within-task ROC-AUC is computed among the errors of each task and averaged with equal task weight. Pooled ROC-AUC is computed over all errors at once. Point estimates with 95\% percentile confidence intervals from 10{,}000 bootstrap resamples of whole tasks, shared across rows.}\label{tab:sup_llm_source_prediction}
  \scriptsize
  \setlength{\tabcolsep}{2pt}
  \begin{tabular}{@{}lcc@{}}
    \toprule
     & \textbf{Within-task ROC-AUC} & \textbf{Pooled ROC-AUC} \\
    \midrule
    L2 logistic regression & 0.69 [0.63,0.76] & 0.64 [0.59,0.70] \\
    EBM & 0.68 [0.61,0.74] & 0.68 [0.62,0.74] \\
    Hist.\ gradient boosting & 0.65 [0.57,0.73] & 0.68 [0.62,0.74] \\
    \bottomrule
  \end{tabular}
\end{table}

%% file: tables/sup_llm_source_prediction_by_config.tex
\begin{table}[!ht]
  \centering
  \caption{\textbf{Human-versus-LLM source prediction for each LLM configuration (\S\ref{sec:methods-source-prediction}).} Each row reports classifiers trained on primitive delta vectors to distinguish human errors from those of a single configuration, with the same features, weighting scheme, task folds and evaluation tasks as Supplementary Table~\ref{tab:sup_llm_source_prediction}. Train includes the configuration's modeled errors on all tasks, and Eval includes those on the 56 matched tasks, on which out-of-fold within-task ROC-AUC is computed. Every row shares the same 651 human training errors and 144 human evaluation errors. Acc. is mean single-attempt accuracy (humans: 59.90\%). Accuracy, reasoning effort and the number of training errors vary together across configurations, so differences between rows should not be attributed to accuracy alone. Point estimates with 95\% percentile confidence intervals from 10{,}000 bootstrap resamples of whole tasks.}\label{tab:sup_llm_source_prediction_by_config}
  \scriptsize
  \setlength{\tabcolsep}{2pt}
  \begin{tabular}{@{}lcccccc@{}}
    \toprule
    & & \multicolumn{2}{c}{\textbf{LLM errors}} & \multicolumn{3}{c}{\textbf{Within-task ROC-AUC}} \\
    \cmidrule(lr){3-4}\cmidrule(lr){5-7}
     & \textbf{Acc.\ (\%)} & \textbf{Train} & \textbf{Eval.} & \textbf{Logistic L2} & \textbf{EBM} & \textbf{Grad.\ boosting} \\
    \midrule
    \textbf{Qwen3.5-27B} & 80.63 & 197 & 112 & 0.61 [0.52,0.69] & 0.59 [0.49,0.69] & 0.58 [0.48,0.67] \\
    \textbf{GPT-OSS (high)} & 72.28 & 236 & 128 & 0.60 [0.51,0.69] & 0.57 [0.47,0.66] & 0.63 [0.53,0.72] \\
    \textbf{GPT-OSS (mid)} & 50.96 & 714 & 221 & 0.68 [0.59,0.76] & 0.69 [0.61,0.76] & 0.68 [0.60,0.76] \\
    \textbf{GPT-OSS (low)} & 25.02 & 991 & 206 & 0.74 [0.67,0.82] & 0.77 [0.69,0.84] & 0.77 [0.68,0.84] \\
    \bottomrule
  \end{tabular}
\end{table}